\documentclass{article}

\usepackage{microtype}
\usepackage{graphicx}
\usepackage{subcaption}
\usepackage{booktabs} % for professional tables

\usepackage{hyperref}

\usepackage[accepted]{icml2026}

\usepackage{amsmath}
\usepackage{amssymb}
\usepackage{mathtools}
\usepackage{amsthm}

\usepackage[capitalize, noabbrev]{cleveref}

\usepackage{tcolorbox}

\theoremstyle{plain}

\theoremstyle{definition}

\theoremstyle{remark}

\usepackage[textsize=tiny]{todonotes}

\icmltitlerunning{The Shape of Addition: Geometric Structures of Arithmetic in Large Language Models}

\begin{document}
  \twocolumn[ \icmltitle{The Shape of Addition: Geometric Structures of Arithmetic\\ in Large Language Models}

  % It is OKAY to include author information, even for blind submissions: the
  % style file will automatically remove it for you unless you've provided
  % the [accepted] option to the icml2026 package.

  % List of affiliations: The first argument should be a (short) identifier you
  % will use later to specify author affiliations Academic affiliations
  % should list Department, University, City, Region, Country Industry
  % affiliations should list Company, City, Region, Country

  % You can specify symbols, otherwise they are numbered in order. Ideally, you
  % should not use this facility. Affiliations will be numbered in order of
  % appearance and this is the preferred way.
  \icmlsetsymbol{equal}{\dag}
  \icmlsetsymbol{corrauthor}{*}

  \begin{icmlauthorlist}
    \icmlauthor{Liuyuan Wen}{equal,inst1} \icmlauthor{Xun Zhu}{equal,inst1} \icmlauthor{Lihao Huang}{equal,inst1}
    \icmlauthor{Wenbin Li}{corrauthor,inst1} \icmlauthor{Yang Gao}{inst1}
  \end{icmlauthorlist}

  % \icmlaffiliation{inst1}{School of Intelligence Science and Technology, Nanjing University, Suzhou Campus, Suzhou, Jiangsu, China}
  \icmlaffiliation{inst1}{State Key Laboratory of Novel Software Technology, Nanjing University, Nanjing 210023, China}

  \icmlcorrespondingauthor{Wenbin Li}{liwenbin@nju.edu.cn}

  % You may provide any keywords that you find helpful for describing your
  % paper; these are used to populate the "keywords" metadata in the PDF but
  % will not be shown in the document
  \icmlkeywords{Mechanistic Interpretability, Large Language Models, Arithmetic Reasoning, Probing, Representation Geometry, Inference-time Intervention}

  \vskip 0.3in ]

  % this must go after the closing bracket ] following \twocolumn[ ...

  % This command actually creates the footnote in the first column listing the
  % affiliations and the copyright notice. The command takes one argument, which
  % is text to display at the start of the footnote. The \icmlEqualContribution
  % command is standard text for equal contribution. Remove it (just {}) if you
  % do not need this facility.

  % Use ONE of the following lines. DO NOT remove the command.
  % If you have no special notice, KEEP empty braces:
  % \printAffiliationsAndNotice{}  % no special notice (required even if empty)
  % Or, if applicable, use the standard equal contribution text:
  \renewcommand{\icmlEqualContribution}{\textsuperscript{\dag}Equal contribution }
  \printAffiliationsAndNotice{\icmlEqualContribution\textsuperscript{*}Corresponding author }

  \begin{abstract}
    Large Language Models exhibit paradoxical fragility in fundamental arithmetic,
    implying a disconnect between internal computation and discrete output. By analyzing
    the residual stream geometry during multi-operand addition, we identify the
    Iso-Raw-Sum Trajectory (IRST), a geometric structure where representations are
    anchored by semantic digits and modulated by continuous carry fibers. We
    propose the Noisy Quantization Model to explain this geometry, framing arithmetic
    errors as Geometric Slippages caused by internal neural noise pushing a continuous,
    latent Carry Potential across quantization thresholds. This geometric framework
    further elucidates Probe Versatility, explaining how lightweight probes can
    disentangle coexisting latent signals (such as ground truth versus hallucination)
    from a single activation vector. Finally, we validate these insights through
    a geometric consistency check method that effectively detects and corrects
    these quantization failures during inference. Our code is available at \href{https://github.com/RL-MIND/Shape-of-Addition}{https://github.com/RL-MIND/Shape-of-Addition}.
  \end{abstract}

  \section{Introduction}

  Large Language Models (LLMs) have demonstrated remarkable capabilities in
  complex mathematical reasoning, achieving high performance on benchmarks such as
  GSM8K \cite{Cobbe21GSM8K} and MATH \cite{Hendrycks21MATH}. However, a paradox remains:
  despite their proficiency in high-level problem solving, LLMs exhibit surprising
  fragility in fundamental algorithmic primitives \citep{li2025exposing}, particularly
  multi-digit arithmetic \cite{Nanda23Arithmetic}. Even sophisticated models
  frequently commit ``off-by-one'' errors as the number of operands increases, suggesting
  a fundamental disconnect between their internal computation and final discrete
  output.
  While prior research has attempted to model these behaviors through either
  symbolic manipulation \cite{quirke2024arithmetic} or geometric representations
  \cite{Nanda23Arithmetic,kantamneni2025language}, the precise mechanism linking
  internal activation geometries to specific output failures remains elusive.

  Crucially, recent probing studies have deepened this mystery
  \cite{yan2025addition,Baeumel25Lookahead}. Lightweight probes can now accurately
  detect error signals and even decode the ground truth from the residual stream
  of a failing model \cite{su2024unsupervised,Sun25Probing}. This implies a
  critical disconnect: the model internally encodes the correct information, yet
  fails to output it. While existing probes serve as effective diagnostic tools,
  they identify \textit{when} a model errs without explaining \textit{how}
  representation geometry mechanistically induces specific failures like carry
  propagation errors.

  \begin{figure*}[ht]
    % \vskip 0.2in
    \centering
    \centerline{\includegraphics[width=\textwidth]{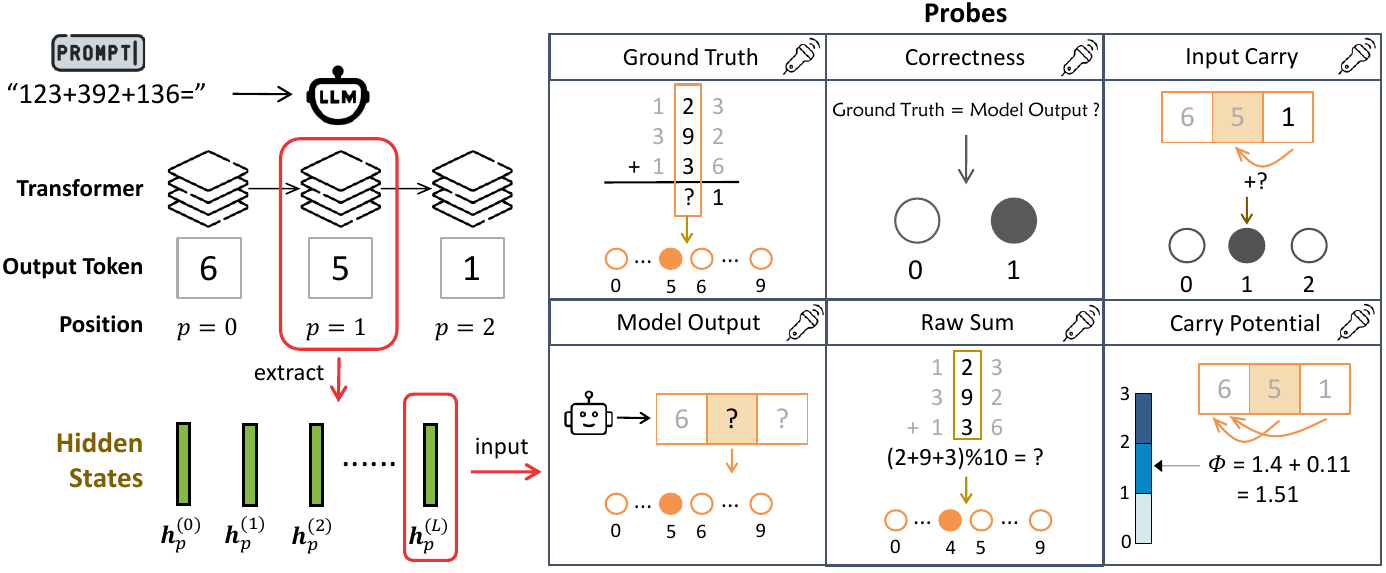}}
    \caption{ \textbf{Overview of our probing framework.} \textbf{(Left)} The LLM
    performs multi-operand addition (e.g., $123+392+136$) in an autoregressive manner.
    At each generation step (e.g., $p=1$, corresponding to the tens digit), we
    extract the hidden state vectors $\boldsymbol{h}_{p}^{(l)}$ (mainly focusing
    on the final layer $L$). \textbf{(Right)} We train versatile probes on these
    activation vectors to decode several critical arithmetic variables, including
    discrete states (Ground Truth, Model Output, Correctness, Input Carry, Raw
    Sum) and continuous variables (Carry Potential $\Phi$, defined in \cref{sec:formal_modeling}).
    The ability to decode these diverse and interrelated signals from a single vector
    motivates our geometric analysis of the residual stream. }
    \label{fig:probes} \vskip -0.1in
  \end{figure*}

  In this work, we investigate the fine-grained internal representations of LLMs
  performing a challenging arithmetic task: the addition of three (or more) 10-digit
  integers. Unlike previous studies that focus on the first token
  \cite{Sun25Probing}, we analyze the residual stream activations at every generated
  digit position (as shown in \cref{fig:probes}).
  Our probing experiments reveal a phenomenon we term \textit{probe versatility}.
  We find that lightweight probes (e.g., logistic regression, MLPs) can simultaneously
  decode diverse signals—including the Ground Truth, Model Output, Input Carry and
  so on—from a single activation vector.
  The coexistence of these signals implies that the representation
  space is actually a highly structured manifold. This raises a fundamental representational
  question: \textit{What is the geometric structure underlying LLM arithmetic representations that allows ground truth and hallucination to coexist?}

\begin{figure*}[!t]
    \centering % 替换 \begin{center}...\end{center}，去除多余垂直间距
    \begin{minipage}[b]{0.535\textwidth}
      \includegraphics[width=\linewidth]{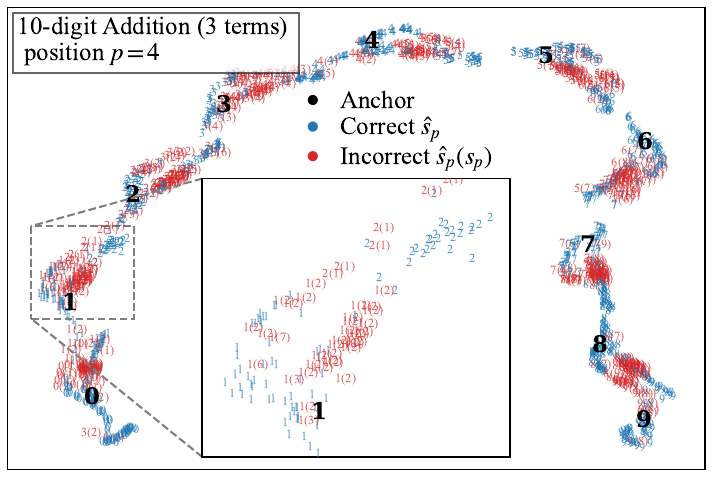}
    \end{minipage}
    \hfill % 自动撑开间距
    \begin{minipage}[b]{0.45\textwidth}
      \includegraphics[width=\linewidth]{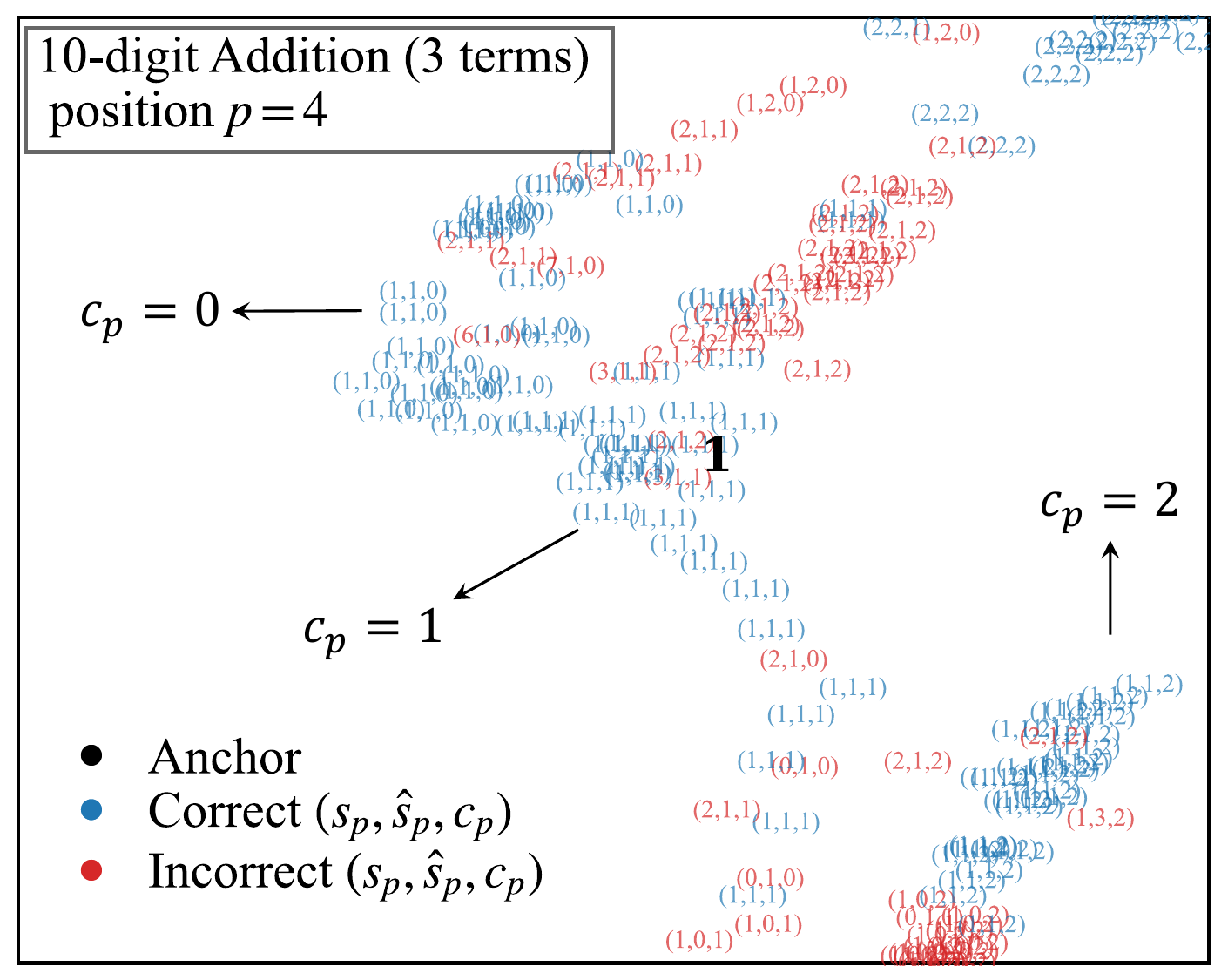}
    \end{minipage}
    \caption{ \textbf{2D UMAP visualization of the arithmetic manifold.} \textbf{(Left)
    Macroscopic Backbone:} Global geometry of $\boldsymbol{h}_{p}^{(L)}$ ($p=4$)
    organized around digit Anchors (0--9). Blue points denote correct samples labeled
    as $\hat{s}_{p}$; red points denote errors labeled as $\hat{s}_{p}(s_{p})$. The
    inset highlights high-error transition zones between digit basins. \textbf{(Right)
    Microscopic Texture:} Magnified view around Anchor 1 labeled with $(s_{p}, \hat
    {s}_{p}, c_{p})$, showing that representations within a digit basin are
    further stratified into distinct fibers by the input carry state $c_{p}$. (The
    geometric consistency in high-dimensional space is validated in Appendix
    \ref{app:geometric_validation}.) }
    \label{fig:umap1} \vskip -0.1in
  \end{figure*}

  To decipher this structure, we employ UMAP \cite{mcinnes2018umap} as dimensionality
  reduction to visualize the internal activation manifolds, primarily at the
  final layer (\cref{fig:umap1}). We discover that representations are organized
  into \textbf{Iso-Raw-Sum Trajectories (IRSTs)}—continuous fibers of constant raw
  sum that thread through a hierarchical geometry composed of a macroscopic backbone
  of digit basins and a microscopic texture of input carry states. Within this
  framework, arithmetic errors are characterizable as \textit{geometric
  slippages}—continuous drifts along an IRST that push the representation across
  the decision boundaries of adjacent digit basins due to ambiguity in carry representation.

  Building on these observations, we propose the \textbf{Noisy Quantization
  Model}, positing that LLMs estimate a continuous \textit{Carry Potential} which
  is subsequently discretized. This framework characterizes arithmetic failures
  as noise-induced threshold crossings, accurately predicting the periodic \textit{bathtub}
  error distribution observed in our experiments (\cref{fig:buthtub1}). Furthermore,
  this geometric perspective helps us demonstrate that the \textit{probe
  versatility} phenomenon is a direct function of the manifold's topological separability
  rather than mere information extractability.

  Finally, we validate this framework through an inference-time self-correction
  method via dual-stream consistency. By enforcing logical alignment between decoded
  local (Raw Sum) and global (Carry Potential) signals, we significantly recover
  performance. This confirms that the model’s internal representations retain the
  correct mathematical components even when the final token selection is erroneous.

  In summary, the main contributions of this work are:
  \begin{itemize}
    \item We identify the \textbf{Iso-Raw-Sum Trajectory (IRST)}, revealing how LLMs
      represent arithmetic states through a hierarchy of raw sum fibers and digit
      basins.

    \item We propose the \textbf{Noisy Quantization Model}, which mechanistically
      models arithmetic errors as quantization failures of a continuous \textit{Carry
      Potential}.

    \item We provide a geometric elucidation of probing phenomenon, showing that
      probing performance hierarchy is structurally determined by the manifold's
      topology.

    \item We introduce a dual-stream consistency check method for inference-time
      intervention, validating the retention of correct latent signals even in
      erroneous cases.
  \end{itemize}

  \section{Preliminaries}
  We investigate the internal mechanisms of LLMs during multi-operand addition.
  This section establishes the mathematical definitions and the probing
  framework used to infer latent internal states, as illustrated in Figure \ref{fig:probes}.

  \subsection{Mathematical Framework}
  We consider the task of adding $n$ integers, denoted as
  $A_{0}, A_{1}, \dots, A_{n-1}$. Each integer $A_{i}$ consists of $m$ decimal digits.
  Let $a_{i,k}\in \{0, 1, \dots, 9\}$ denote the digit of $A_{i}$ at the
  $10^{k}$ position (where the units place corresponds to $k=0$). Therefore,
  each addend can be expressed as $A_{i}= \sum_{k=0}^{m-1}a_{i,k}\cdot 10^{k}$. Let
  $S$ be the sum of these integers, i.e., $S = \sum_{i=0}^{n-1}A_{i}$.

  We construct a dataset $\mathcal{D}$ of $N$ addition problems. The LLM generates
  the sum token-by-token in an autoregressive manner from the most significant digit
  to the least. We define the generation step as the \textit{position} $p$. If
  the target sum $S$ has a length of $P$ digits, the generation position $p \in \{
  0, \dots, P-1\}$ corresponds to the mathematical significance $k = P - 1 - p$.

  Crucially, the arithmetic state at any position $p$ is governed by three
  fundamental variables: the Raw Sum $r_{p}= \sum_{i=0}^{n-1}a_{i, p}$, which is
  the column-wise sum of input digits; the Input Carry $c_{p}$, propagated from
  the lower-order position; and the Ground Truth Digit $s_{p}$. These satisfy
  the modular relation $s_{p}\equiv (r_{p}+ c_{p}) \pmod{10}$.

  To characterize the model's internal state, we introduce parallel variables for
  the model's behavior. Let $\hat{s}_{p}$ denote the Predicted Digit generated
  by the model. We further postulate two latent variables: the Predicted Raw Sum
  $\hat{r}_{p}$ and the Predicted Input Carry $\hat{c}_{p}$, representing the internal
  information the model utilizes to generate $\hat{s}_{p}$. We aggregate these
  into a state tuple for each position:
  \begin{equation}
    \mathcal{X}_{p}= (s_{p}, \hat{s}_{p}, c_{p}, \hat{c}_{p}, r_{p}, \hat{r}_{p})
    .
  \end{equation}
  This formalism allows us to distinguish between the mathematical truths ($s_{p}
  , c_{p}, r_{p}$) and the model's internal representations ($\hat{s}_{p}, \hat{c}
  _{p}, \hat{r}_{p}$).

  \subsection{Internal Activations and Probing}
  For every problem in the dataset $\mathcal{D}$ and at each generation position
  $p$, we extract the internal hidden states $\boldsymbol{h}_{p}^{(l)}$ from
  layer $l$ (primarily focusing on the final layer $L$, layer-wise analysis is
  provided in Appendix \ref{app:probe}). To decode the information encoded in
  these representations, we train lightweight probes on $\boldsymbol{h}_{p}^{(L)}$.
  We target the discrete variables defined in the previous section: the ground truth
  digit $s_{p}$, the predicted digit $\hat{s}_{p}$, the input carry $c_{p}$, and
  the raw sum $r_{p}$. Additionally, we investigate continuous variables such as
  the \textit{Carry Potential} $\Phi$, a scalar value modeling the accumulation
  of arithmetic value from the context, which is defined in \cref{sec:formal_modeling}.

  \subsection{Latent Carry States}
  While $s_{p}$, $\hat{s}_{p}$, $c_{p}$, and $r_{p}$ are directly observable, the
  internally predicted $\hat{c}_{p}$ and $\hat{r}_{p}$ are latent and must be inferred.
  We assume that the model generally performs the local addition operation
  correctly (i.e., $\hat{r}_{p}\approx r_{p}$), see Appendix \ref{app:raw_sum_validation}
  for a validation. So that errors in $\hat{s}_{p}$ primarily stem from
  incorrect carry states. By analyzing the modular arithmetic relations, we can
  map the observed output errors to internal carry deviations. Specifically, if
  the model underestimates the digit by 1, it implies a leakage in the internal carry
  ($\hat{c}_{p}= c_{p}- 1$); conversely, if it overestimates by 1, it implies a
  hallucinated carry ($\hat{c}_{p}= c_{p}+ 1$). A detailed derivation is provided
  in Appendix \ref{app:latent_carry}.

  \begin{figure*}[ht]
    \centering % 替换 \begin{center}...\end{center}，去除多余垂直间距
    \begin{minipage}[b]{0.49\textwidth}
      \includegraphics[width=\linewidth]{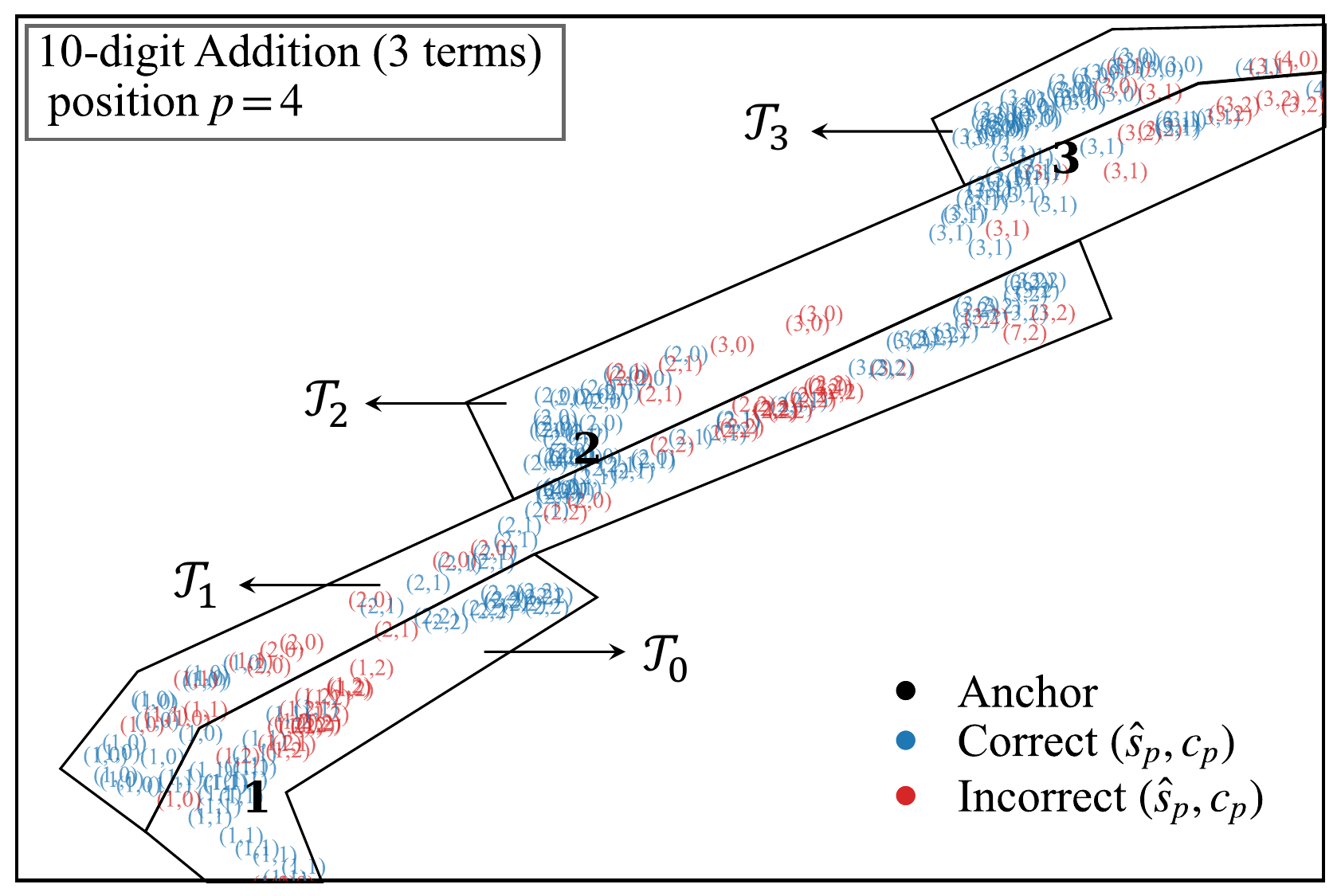}
    \end{minipage}
    \hfill % 自动撑开间距
    \begin{minipage}[b]{0.49\textwidth}
      \includegraphics[width=\linewidth]{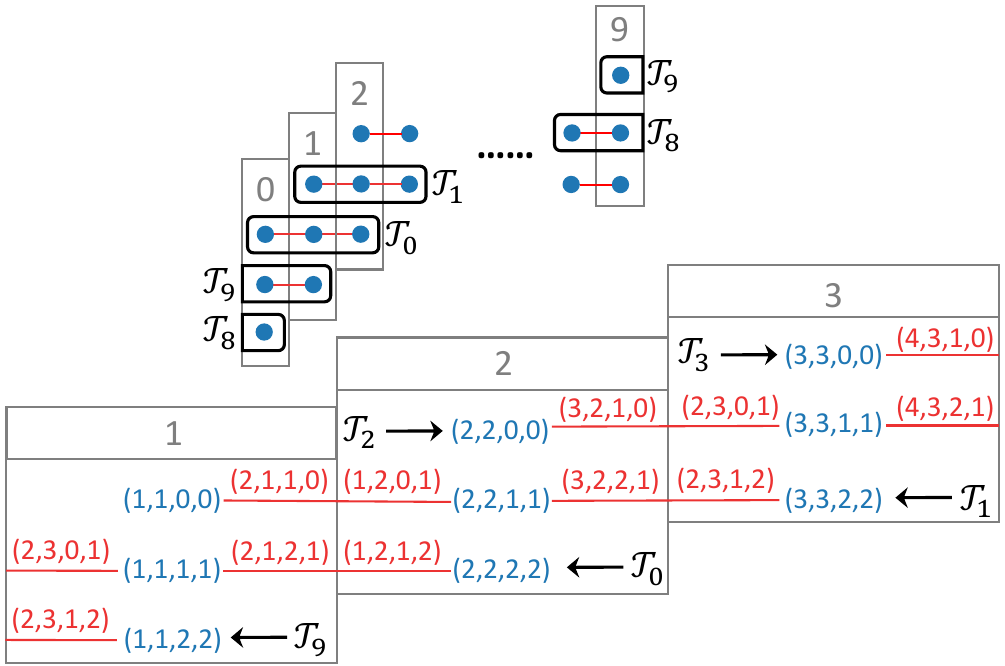}
    \end{minipage}
    \caption{ \textbf{The Iso-Raw-Sum Trajectory (IRST) framework of the
    arithmetic manifold.} \textbf{(Left)} Magnified UMAP projection around digit
    Anchors 1, 2, and 3. Points are labeled with $(\hat{s}_{p}, c_{p})$. The
    geometry reveals distinct IRSTs ($\mathcal{T}_{0}\sim \mathcal{T}_{3}$) that
    act as continuous ``threads'' piercing through adjacent digit basins. For instance,
    $\mathcal{T}_{1}$ (where $r_{p}\bmod 10 = 1$) connects stable nodes
    $(1,0) \leftrightarrow (2,1) \leftrightarrow (3,2)$ as the input carry
    increases. \textbf{(Top-Right)} Abstract representation of the global
    manifold, illustrating how parallel raw-sum fibers $(\mathcal{T}_{r})$ organize
    the representation space across all digit basins (0--9). \textbf{(Bottom-Right)}
    Detailed state-level schematic of the IRST framework. Gray boxes represent digit
    basins; states are defined by the tuple
    $(s_{p}, \hat{s}_{p}, c_{p}, \hat{c}_{p})$. Blue denotes correct predictions,
    while red denotes \textit{geometric slippages} (errors). These errors are
    concentrated at transition zones between basins along an IRST, resulting from
    the model's miscalculation of the input carry ($\hat{c}_{p}= c_{p}\pm 1$). }
    \label{fig:irst_analysis} \vskip -0.1in
  \end{figure*}

  \section{Representational Geometry}
  \label{sec:represent}

  \subsection{Experimental Setup}
  To empirically investigate the latent structure of arithmetic reasoning, we
  conduct experiments using the Qwen3-4B model \cite{yang2025qwen3}, which consists
  of $L=36$ transformer layers. We construct a dataset $\mathcal{D}$ containing
  $N=10,000$ addition problems, each involving three $10$-digit integers. In
  this $3$-term addition setup, the input carry $c_{p}$ at any position can take
  values in $\{0, 1, 2\}$. We focus our analysis on the hidden states
  $\boldsymbol{h}_{p}^{(L)}$ extracted from the final layer at position $p=4$\footnote{We
  select $p=4$ as a representative middle position.}. 
  Analysis of other
  interior positions yields qualitatively similar findings, while boundary positions
  are discussed in Appendix \ref{app:noise_scaling}.

  We visualize the high-dimensional activation space using UMAP \cite{mcinnes2018umap}
  dimensionality reduction with \textit{cosine} distance as the distance metric
  (see Appendix \ref{app:dr_robustness} for other reduction methods). To provide
  a semantic reference, we include the output unembedding vectors of the digit
  tokens $0$ through $9$ in the reduction process, which serve as fixed \textit{Anchors}\footnote{These
  anchors serve exclusively as auxiliary landmarks for semantic grounding; their
  inclusion does not alter the intrinsic geometry of the activation vectors
  in the UMAP projection.} in the resulting 2D map. The visualization, presented
  in \cref{fig:umap1}, reveals a highly organized geometric manifold that
  encodes the model's arithmetic state through a hierarchical geometric
  structure. More detailed experimental settings are provided in Appendix \ref{app:exp_details}.

  \subsection{Geometric Structure}

  The geometry of the last-layer activations exhibits a clear separation between
  global semantic identity and local arithmetic state. We characterize this structure
  as having a macroscopic \textit{backbone} modulated by a microscopic \textit{texture}.
  The macroscopic backbone is dominated by the identity of the output tokens. As
  seen in the left panel of \cref{fig:umap1}, the activation vectors are
  strongly clustered into ten distinct basins, each centered around a specific digit
  anchor. This indicates that the primary organization of the residual stream at
  the final layer is determined by the categorical selection of the next token.

  Within each digit basin, the representation space possesses a fine-grained microscopic
  texture. As illustrated in the magnified view around Anchor 1 in the right
  panel of \cref{fig:umap1}, the data points are not distributed uniformly.
  Instead, they are stratified into parallel sub-manifolds or \textit{fibers}
  determined by the ground truth input carry $c_{p}$. Specifically, for any
  digit basin, we observe three distinct clusters corresponding to the states $c_{p}
  \in \{0, 1, 2\}$. This suggests that the model internalizes the underlying
  algebraic structure \cite{chang2024unraveling} to distinguish the source of a
  digit. For instance, differentiating ``a 3 derived from no carry'' from ``a 3
  derived from a carry of 1''.

  The distribution of correct (blue) and incorrect (red) samples further
  elucidates the reliability of this manifold in \cref{fig:umap1}. Correct predictions
  cluster densely at the center of these carry-specific fibers, representing
  stable states of computation. Conversely, errors predominantly occur in the
  sparse transition zones between digit basins or between carry fibers. As shown
  in the inset of the left panel, incorrect samples often lie on the fringes of
  their ground truth digit basins, physically drifting toward adjacent anchors.
  This drift manifests empirically as the prevalent ``off-by-one'' errors—a
  common failure mode cataloged in mathematical reasoning benchmarks
  \cite{zhang2025mathematical}, confirming that geometric adjacency on the manifold
  aligns with numerical adjacency.
  This spatial distribution suggests that arithmetic failures are essentially
  geometric instabilities, where \textit{the internal representation fails to
  align with the correct geometric trajectory.}

  \section{Geometric Analysis of Internal States}

  Building on the UMAP visualization of the last-layer activations, we propose a
  geometric framework called the \textbf{Iso-Raw-Sum Trajectory (IRST)} to
  interpret how the LLM encodes arithmetic states (\cref{fig:irst_analysis}).

  \subsection{Iso-Raw-Sum Trajectories (IRST)}

  The key insight connecting the discrete digit clusters is the conservation of the
  \textit{raw sum} $r_{p}$. We define an \textbf{IRST}, denoted as $\mathcal{T}_{r}$,
  as a continuous manifold connecting all internal states that share the same
  raw sum $r_{p}= r$. Recalling the identity $\hat{s}_{p}= (r_{p}+ \hat{c}_{p}) \pmod
  {10}$, a trajectory $\mathcal{T}_{r}$ does not stay within a single digit
  basin but traverses across adjacent basins as the carry $\hat{c}_{p}$ varies.

  As illustrated in \cref{fig:irst_analysis} (Left), consider the trajectory $\mathcal{T}
  _{1}$ (where raw sum $r_{p}=1$). It acts as a continuous \textit{thread} that pierces
  through the boundaries of Anchors 1, 2, and 3, connecting three stable fixed
  nodes in a linear sequence:
  \begin{equation}
    \mathcal{T}_{1}: \underbrace{(1,1,0,0)}_{\text{Anchor 1, } c=0}\leftrightarrow
    \underbrace{(2,2,1,1)}_{\text{Anchor 2, } c=1}\leftrightarrow \underbrace{(3,3,2,2)}
    _{\text{Anchor 3, } c=2}.
  \end{equation}
  Here, we use the tuple notation $(s_{p}, \hat{s}_{p}, c_{p}, \hat{c}_{p})$. The
  model transitions between output digits $s$ and $s+1$ by sliding along these fixed
  raw sum fibers, driven by the shift in the internal carry state.

  \subsection{Geometric Slippage on Trajectories}

  In this framework, arithmetic errors are not random noise but specific \textit{Geometric
  Slippages} along an IRST. They occur when activation vectors drift onto
  unstable segments connecting stable nodes, rather than aligning with the stable
  nodes themselves.

  Focusing on the transition between Anchor 1 and Anchor 2 (see
  \cref{fig:irst_analysis}, Bottom-Right), we identify two distinct error paths corresponding
  to different raw sums:

  \paragraph{Scenario 1: Slippage on $\mathcal{T}_{1}$ (Raw Sum = 1).}
  This trajectory bridges the stable node $(1,1,0,0)$ and $(2,2,1,1)$. Errors found
  on the segment connecting these nodes represent a specific confusion between carry
  0 and carry 1. In the case of \textit{Overestimation (Hallucination)}, characterized
  by state $(1,2,0,1)$, the ground truth requires $c=0$, but the model hallucinates
  $\hat{c}=1$, geometrically pushing the representation from Anchor 1 toward
  Anchor 2. Conversely, \textit{Underestimation (Leakage)} corresponds to state
  $(2,1,1,0)$; here, the ground truth involves $c=1$, but the model misses the carry
  ($\hat{c}=0$), causing the vector to slide back from the correct basin of
  Anchor 2 into Anchor 1.

  \paragraph{Scenario 2: Slippage on $\mathcal{T}_{0}$ (Raw Sum = 0).}
  Similarly, $\mathcal{T}_{0}$ bridges $(1,1,1,1)$ and $(2,2,2,2)$, involving
  confusion between carry 1 and 2. Here, hallucination manifests as $(1,2,1,2)$,
  where the model erroneously jumps to $\hat{c}=2$ despite $c=1$. Leakage appears
  as $(2,1,2,1)$, where the model fails to sustain the carry, retreating to the lower
  state $\hat{c}=1$.

  The geometric nature of arithmetic errors is therefore defined by the proximity
  to the decision boundary between anchors along a specific IRST. Correct
  predictions cluster densely around the stable nodes, while errors are distributed
  along the sparse connecting paths. In these transitional regions, the unembedding
  layer struggles to distinguish between the two digits, as the activation
  contains mixed signals: the raw sum suggests continuity, but the ambiguous
  carry state places the vector in a ``no-man's-land'' between two semantic
  truths.

  \begin{figure}[!t]
    % \vskip 0.2in
    \centering
    \centerline{\includegraphics[width=\columnwidth]{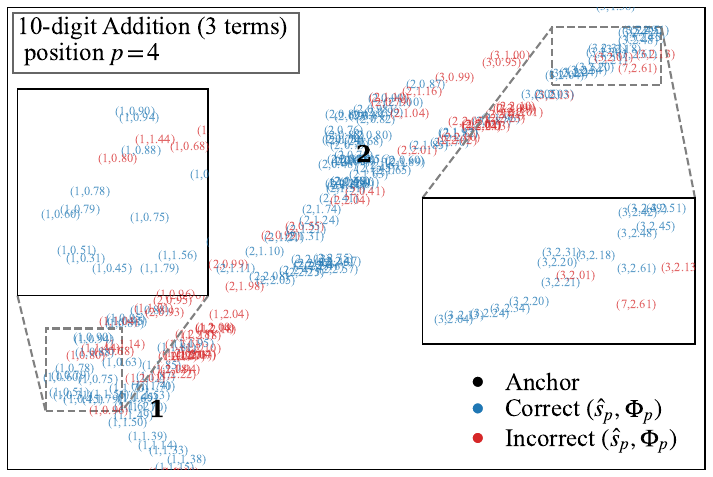}}
    \caption{ \textbf{Geometric unfolding of the IRST by Carry Potential.}
    Visualization of the trajectory $\mathcal{T}_{1}$, where each point is labeled
    with the tuple $(\hat{s}_{p}, \Phi_{p})$. \textbf{(Insets)} The geometry correlates
    strongly with the continuous value of the theoretical carry potential $\Phi_{p}$.
    The left inset highlights the region where $\Phi_{p}< 1$ (approaching the threshold
    from below), corresponding to the stable node of $c_{p}=0$. The right inset
    displays the region where $\Phi_{p}> 2$, corresponding to the node of $c_{p}=
    2$. The smooth spatial progression of $\Phi_{p}$ values along the manifold
    confirms that the IRST is physically organized by the continuous \textit{thrust}
    to carry, validating the continuity assumption of the Noisy Quantization
    Model. }
    \label{fig:potential_umap} \vskip -0.1in
  \end{figure}

  \section{Formal Modeling of Carry Dynamics}
  \label{sec:formal_modeling}

  Our geometric analysis reveals that the IRSTs act as continuous manifolds
  connecting discrete digit anchors. This continuity suggests that the LLM's internal
  representation of arithmetic is not purely symbolic. Instead, we hypothesize that
  the model maintains a continuous, latent variable representing the ``pressure''
  to carry, which is subsequently discretized. In this section, we verify this hypothesis
  by proposing the \textbf{Noisy Quantization Model}. We demonstrate that arithmetic
  errors are essentially predictable consequences of signal detection noise near
  quantization boundaries.

  \subsection{Carry Potential ($\Phi$)}
  \label{subsec:carry_potential} We define the \textit{Carry Potential}, denoted
  as $\Phi_{p}\in \mathbb{R}_{\geq 0}$, as the theoretical accumulation of
  numerical value flowing from lower-order positions (the right context) into
  the current position $p$. Unlike the discrete ground truth carry $c_{p}\in \mathbb{Z}$,
  the potential $\Phi_{p}$ is a continuous scalar derived from the weighted sum
  of raw sums in the context:
  \begin{equation}
    \Phi_{p}= \sum_{j=1}^{P-1-p}\frac{r_{p+j}}{10^{j}},
  \end{equation}
  where $r_{p+j}$ is the raw sum of the digits at relative position $j$ to the right
  of $p$. Physically, $\Phi_{p}$ represents the \textit{thrust} or \textit{momentum}
  of the carry signal, and the discrete ground truth carry corresponds to its
  integer floor: $c_{p}= \lfloor \Phi_{p}\rfloor$. For example, as illustrated in
  \cref{fig:probes}, the rightward context yields a potential of
  $\Phi = 1.4 + 0.11 = 1.51$. This continuous value physically represents the \textit{thrust}
  to carry, which is subsequently quantized to the discrete carry $c_{p}= \lfloor
  1.51 \rfloor = 1$.

  This theoretical definition is strongly corroborated by our empirical observations.
  As shown in \cref{fig:potential_umap}, when we label the activation manifold with
  the calculated values of $\Phi_{p}$, we observe a smooth spatial gradient
  along the IRST. The representation explicitly encodes the continuous magnitude
  of $\Phi_{p}$, placing states with $\Phi_{p}\approx 0.9$ (high risk of carry)
  spatially closer to the next carry basin than states with
  $\Phi_{p}\approx 0.1$.

  \subsection{Noisy Quantization Hypothesis}
  We posit that the LLM estimates this external potential $\Phi_{p}$ via its
  internal representations. We further hypothesize that, due in part to the dilution
  of attention over long contexts and the limited precision of superposition,
  this estimation is corrupted by neural noise. We model the \textit{Perceived
  Carry Potential},
  $\hat{\Phi}_{p}$, as:
  \begin{equation}
    \hat{\Phi}_{p}= \Phi_{p}+ \epsilon, \quad \epsilon \sim \mathcal{N}(0, \sigma
    ^{2}),
  \end{equation}
  where $\epsilon$ represents additive Gaussian noise with zero mean and
  variance $\sigma^{2}$. The parameter $\sigma$, which we term the \textit{Cognitive
  Noise Level}, encapsulates the effective uncertainty in the model's estimation
  imposed by the current task complexity. The model then generates the discrete predicted
  input carry $\hat{c}_{p}$ by performing a quantization operation (flooring) on
  this noisy signal:
  \begin{equation}
    \hat{c}_{p}= \lfloor \hat{\Phi}_{p}\rfloor = \lfloor \Phi_{p}+ \epsilon \rfloor
    .
  \end{equation}
  Errors occur when the noise $\epsilon$ is sufficient to push the perceived
  potential $\hat{\Phi}_{p}$ across an integer quantization boundary.

  \begin{figure}[t]
    % \vskip 0.2in
    \centering
    \begin{minipage}[t]{\columnwidth}
      \centering
      \includegraphics[width=0.8\linewidth]{
        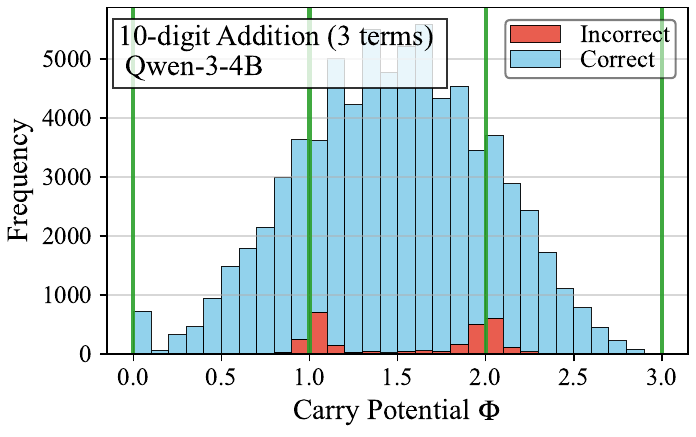
      }
    \end{minipage}
    \hfill % 左右图之间的空白（自动填充）
    \begin{minipage}[t]{\columnwidth}
      \centering
      \includegraphics[width=0.8\linewidth]{
        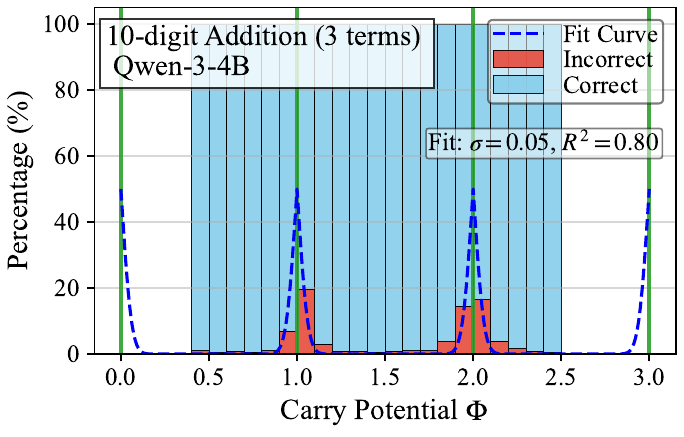
      }
    \end{minipage}
    \caption{ \textbf{Empirical validation of the Noisy Quantization Model.}
    \textbf{(Top)} The distribution of Carry Potential $\Phi$ across all generated
    positions $p$ in the dataset. Green vertical lines indicate integer
    quantization thresholds ($1.0, 2.0, \dots$). \textbf{(Bottom)} The
    conditional error rate as a function of $\Phi$. The empirical data (red bars,
    off-by-one errors only) exhibits a distinct periodic \textit{bathtub} shape,
    spiking near integer boundaries (intervals with low sample density are omitted).
    The theoretical error curve (dashed blue line), derived from
    \cref{eq:double_q}, fits the data with high fidelity ($R^{2}=0.80$),
    estimating the model's internal noise level at $\sigma \approx 0.05$. }
    \label{fig:buthtub1} \vskip -0.1in
  \end{figure}

  \subsection{Bathtub Error Rate}
  This formulation constitutes a threshold detection problem. The likelihood of
  an error depends on the distance of the potential $\Phi_{p}$ to the nearest
  integer boundary.

  Let $\delta(\Phi) = \Phi \pmod 1$ be the fractional part of the potential, the
  total probability of an off-by-one carry error given a potential $\Phi$ is the
  sum of the probabilities of \textit{Underestimation (Leakage)} and \textit{Overestimation
  (Hallucination)}:
  \begin{equation}
    \label{eq:double_q}P(\text{Error}\mid \Phi) = \underbrace{Q\left(\frac{\delta(\Phi)}{\sigma}\right)}
    _{\text{Leakage Risk}}+ \underbrace{Q\left(\frac{1 - \delta(\Phi)}{\sigma}\right)}
    _{\text{Hallucination Risk}},
  \end{equation}
  where $Q(\cdot)$ is the standard Q-function. Detailed derivation provided in Appendix
  \ref{app:error_derivation}. This equation predicts a periodic \textit{bathtub}
  error distribution: error rates spike when $\Phi$ is close to an integer $i$ (\textit{metastability})
  and vanish when $\Phi$ is near $i + 0.5$ (\textit{robust plateaus}).

  To empirically validate this prediction, we computed the analytical carry potential
  $\Phi$ across the entire dataset $\mathcal{D}$. Both the global distribution
  of potentials (\cref{fig:buthtub1}, Top) and the conditional error rate (\cref{fig:buthtub1},
  Bottom) reveals a striking periodic \textit{bathtub} pattern. Error rates
  spike drastically near integer boundaries (e.g., $\Phi \approx 1.0, 2.0$), identifying
  them as regions of \textit{metastability} where the signal is most ambiguous. Conversely,
  the error rates reach their minima at half-integer values (e.g., $\Phi \approx
  1.5$). These \textit{robust plateaus} represent the states where the signal is
  maximally distinct from the quantization thresholds.

  By fitting the theoretical curve to this empirical distribution, we extract a noise
  parameter of $\sigma \approx 0.05$. This value quantifies the model's
  processing uncertainty, reflecting the precision limit of internal signal processing
  specific to this computational context. The close alignment between theory and
  observation confirms that arithmetic errors are mechanistically determined by
  this internal noise pushing the carry potential across quantization boundaries.
  More investigation into how this noise scales with task complexity is provided
  in Appendix \ref{app:noise_scaling}.

  \begin{table}[!t]
    \caption{Probing accuracy on the final layer activations ($p=4$). Datasets are
    individually balanced for each probe target.}
    \label{tab:probe_performance}
    \centering
    \begin{small}
      \begin{sc}
        \begin{tabular}{llc}
          \toprule Target Variable & Symbol                              & Acc              \\
          \midrule Ground Truth    & $s_{p}$                             & 94.85\%          \\
          Model Output             & $\hat{s}_{p}$                       & 98.81\%          \\
          Correctness              & $\mathbb{I}(\hat{s}_{p}\neq s_{p})$ & 82.41\%          \\
          Raw Sum (Mod 10)         & $r_{p}\pmod{10}$                    & 98.60\%          \\
          Input Carry              & $c_{p}$                             & 96.84\%          \\
          Carry Potential          & $\Phi_{p}$                          & 92.08\%(floored) \\
          \bottomrule
        \end{tabular}
      \end{sc}
    \end{small}
    \vskip -0.1in
  \end{table}

  \section{Trajectory-Level Validation and Causal Steering}
  \label{sec:trajectory_validation}

  The IRST geometry illustrated in \Cref{fig:irst_analysis,fig:potential_umap}
  predicts more than a visually plausible 2D embedding: along any IRST,
  arithmetic states should be ordered by a continuous carry coordinate, and
  moving a representation along that coordinate should causally change the
  generated digit. To test this directly, we define a
  trajectory-local steering direction between two adjacent stable centroids
  $\boldsymbol{\mu}_{a}$ and $\boldsymbol{\mu}_{b}$ on the same IRST:
  \begin{equation}
    \vec{v}_{steer}= \frac{\boldsymbol{\mu}_{b}- \boldsymbol{\mu}_{a}}{\|
    \boldsymbol{\mu}_{b}- \boldsymbol{\mu}_{a}\|}.
  \end{equation}
  We then intervene on a final-layer pre-norm activation $\boldsymbol{h}$ by
  \begin{equation}
    \tilde{\boldsymbol{h}}(\alpha) = \boldsymbol{h}+ \alpha \cdot \vec{v}_{steer},
  \end{equation}
  where $\alpha > 0$ moves the state toward the higher-carry basin and
  $\alpha < 0$ suppresses the carry signal. If the carry potential is genuinely
  represented along this direction, samples near a quantization boundary should flip under smaller
  perturbations than deep in-basin states.

\begin{figure*}[!t]
    \centering
    \begin{minipage}[b]{0.47\textwidth}
      \includegraphics[width=0.94\linewidth]{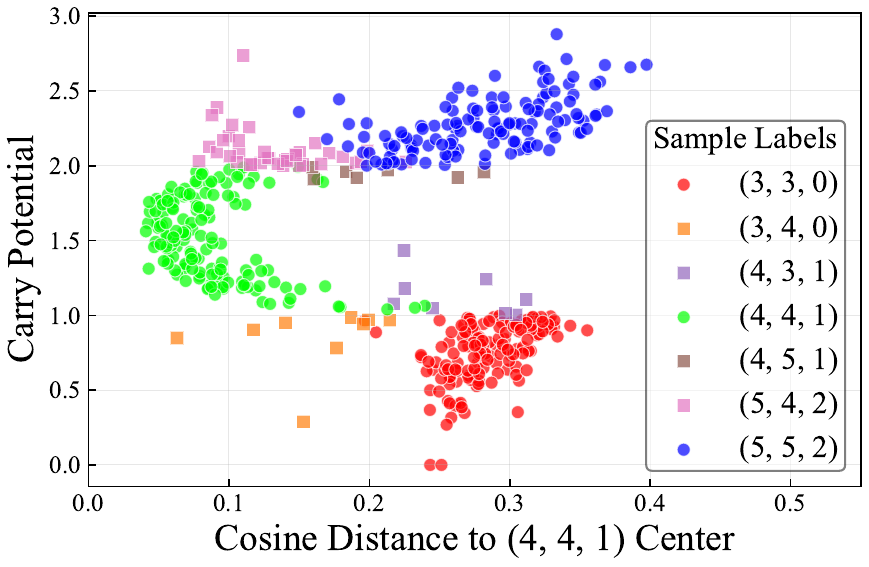}
    \end{minipage}
    \hfill
    \begin{minipage}[b]{0.47\textwidth}
      \includegraphics[width=0.94\linewidth]{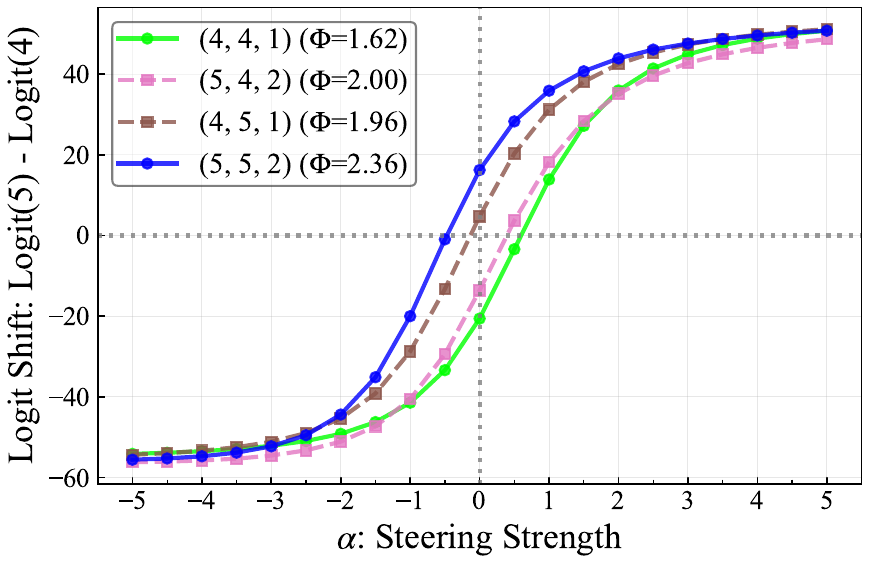}
    \end{minipage}
    \caption{\textbf{Representative trajectory-level validation on $\mathcal{T}_{3}$.}
    The markers are labeled as $(s_{p}, \hat{s}_{p}, c_{p})$. Circles denote
    correct predictions and squares denote errors. \textbf{(Left)} Empirical
    projection of last-layer activations for samples in $\mathcal{T}_{3}$. The
    x-axis shows the cosine distance from the central centroid $(4,4,1)$. The
    manifold exhibits a clear V-shaped progression connecting stable basins.
    Crucially, error states such as $(4,3,1)$ and $(5,4,2)$ cluster in the
    transition regions between stable fibers, validating the \textit{geometric
    slippage} hypothesis. \textbf{(Right)} Causal steering results. We inject
    the carry vector $\vec{v}_{steer}$ (derived from contrastive means) and
    measure the Logit Shift ($\text{Logit}(5)-\text{Logit}(4)$). The results show
    characteristic sigmoid phase transitions: stable states require larger
    $|\alpha|$ to flip than metastable error states near the boundary.}
    \label{fig:app_t3_analysis} \vskip -0.1in
  \end{figure*}

  We evaluate this prediction on a representative trajectory $\mathcal{T}_{3}$ at
  $p=4$, where the adjacent stable states $(3,3,0)$, $(4,4,1)$, and $(5,5,2)$
  are all well populated. For this case, we instantiate the steering vector using
  $\boldsymbol{\mu}_{a}=\boldsymbol{\mu}_{(4,4,1)}$ and
  $\boldsymbol{\mu}_{b}=\boldsymbol{\mu}_{(5,5,2)}$. In \Cref{fig:app_t3_analysis}
  (Left), plotting the analytical Carry Potential $\Phi$ against the cosine
  distance to $\boldsymbol{\mu}_{(4,4,1)}$ yields the predicted V-shaped
  progression, with off-by-one errors such as $(4,5,1)$ and $(5,4,2)$ concentrated
  in the transition regions rather than forming a separate cluster. In
  \Cref{fig:app_t3_analysis} (Right), the boundary-adjacent states $(4,5,1)$ and
  $(5,4,2)$ switch at substantially smaller perturbation magnitudes
  ($\alpha \approx -0.1$ and $\alpha \approx 0.3$, respectively) than the stable
  states $(5,5,2)$ and $(4,4,1)$
  ($\alpha \approx -0.5$ and $\alpha \approx 0.5$). This separation of critical
  thresholds supports the claim that the model's arithmetic decision is governed
  by a carry-like continuous direction. Additional
  analysis are deferred to
  Appendix~\ref{app:geometric_validation}.

  \section{Geometric Origins of Probing Phenomenon}
  \label{sec:probing_interpretation}

  Our discovery offers a unified geometric lens to interpret the varying performance
  of linear probes reported in literature. We trained same logistic regression
  probes on a same balanced datasets at $p=4$. As shown in Table \ref{tab:probe_performance},
  the accuracy differences are essentially direct measures of the geometric
  separability of the underlying manifold.

  \paragraph{Model Output vs. Ground Truth.}
  The probe for Model Output achieves near-perfect accuracy (98.81\%), effectively
  mimicking the unembedding matrix to identify the Voronoi basin of the predicted
  anchor. In contrast, the Ground Truth probe lags significantly (94.85\%). This
  gap confirms our \textit{geometric slippage} hypothesis: Activation vectors physically
  drift to incorrect basins during errors. For the probe to recover the ground
  truth in these cases, it would need to map the incorrect basin back to the correct
  label, which contradicts the local geometry.

  \paragraph{Correctness as Boundary Detection.}
  The Correctness probe yields the lowest accuracy (82.41\%). Geometrically,
  this probe acts as a stability detector, attempting to distinguish samples deep
  within a basin (correct) from those in the transition zones (likely errors).
  The moderate performance implies that the manifold is continuous; there is no sharp
  error flag or discrete boundary separating correct and incorrect states, only
  a smooth gradient of ambiguity along the IRST.

  \paragraph{Decoding the IRST Structure.}
  The high accuracy of the Raw Sum probe (98.60\%) supports the existence of
  distinct IRSTs. Since most arithmetic errors are off-by-one carry shifts, the
  representation slides along the same IRST rather than jumping to a different
  one, preserving the linear separability of the raw sum. Similarly, the Input Carry
  probe (96.84\%) successfully disentangles the parallel fibers (carry 0, 1, 2)
  within these trajectories.

  \paragraph{Continuous Potential vs. Discrete Carry.}
  The Carry Potential probe, trained via regression, achieves 92.08\% accuracy (after
  quantization). This slightly lower performance compared to the discrete carry probe
  reflects the complexity of mapping the continuous position along an IRST to a discrete
  integer. The probe is essentially measuring the distance traveled along the
  trajectory; the fact that it correlates highly with the discrete carry validates
  that the IRST is physically organized by this continuous magnitude.

  \section{Inference-Time Self-Correction}
  \label{sec:method_correction}

  Building on the Noisy Quantization model, we propose a geometric inference-time
  intervention acting as a causal probe to validate the IRST structure. By
  mitigating quantization noise through logical consistency constraints, we show
  that the underlying arithmetic signal remains intact and recoverable.

  \subsection{Dual-Stream Consistency Protocol}

  We interpret the activation vector $\boldsymbol{h}_{p}^{(L)}$ at the final
  layer through two orthogonal lightweight probes. First, we decode the \textit{Local
  Calculation Stream} using a classification probe $f_{\theta_r}$ to obtain the predicted
  raw sum digit $\hat{r}_{p}$. Second, we decode the \textit{Global Context
  Stream} using a regression probe $f_{\theta_{\phi}}$ to estimate the
  continuous carry potential $\hat{\Phi}_{p}$. As demonstrated in \cref{sec:formal_modeling},
  $\hat{\Phi}_{p}$ acts as a proxy for the cumulative context information, while
  $\hat{r}_{p}$ represents the immediate column-wise arithmetic.

  In a consistent arithmetic system, the generated token $\hat{s}_{p}$ must
  satisfy the modular identity $\hat{s}_{p}\equiv (\hat{r}_{p}+ c) \pmod{10}$
  for some valid carry $c$. However, due to the inherent cognitive noise
  $\sigma$ in estimating $\hat{\Phi}_{p}$, enforcing a strict integer carry $c =
  \lfloor \hat{\Phi}_{p}\rfloor$ is prone to false positives near quantization boundaries.
  To address this, we introduce a robust consistency check. We define a set of
  \textit{Plausible Carries} $\mathcal{K}_{p}(\delta)$ derived from the $\delta$-neighborhood
  of the estimated potential:
  
  \begin{equation}
    \mathcal{K}_{p}(\delta) = \left\{ \lfloor \phi \rfloor \mid \phi \in [\hat{\Phi}
    _{p}- \delta, \hat{\Phi}_{p}+ \delta] \right\}.
  \end{equation}
  
  The generation $\hat{s}_{p}$ is deemed consistent if it can be explained by
  the local raw sum $\hat{r}_{p}$ and any plausible carry
  $c \in \mathcal{K}_{p}(\delta)$.

  If the model's output $\hat{s}_{p}$ fails this consistency check, it indicates
  a geometric divergence where the representation has drifted off the valid IRST.
  In such cases, we intervene by overriding the output logits: $\hat{s}_{new}
  = (\hat{r}_{p}+ \lfloor \hat{\Phi}_{p}\rfloor) \pmod{10}$. This effectively
  projects the divergent state back onto the correct manifold fiber.

  \begin{table}[t]
    \caption{Performance comparison of different correction methods on Token-level
    Accuracy, True Positive Correction, and False Positive Preservation.}
    \label{tab:method-comparison}
    \centering
    \begin{small}
      \begin{sc}
        \begin{tabular}{lcccc}
          \toprule Method     & Token Acc        & TP Corr          & FP Pres          \\
          \midrule Original   & 86.26\%          & /                & /                \\
          Re-Prompting        & 79.90\%          & 0.08\%           & \textbf{99.98\%} \\
          Steering            & 88.27\%          & 30.58\%          & 96.97\%          \\
          Replacement         & 89.13\%          & 31.73\%          & 97.65\%          \\
          Ours($\delta=0$)    & 87.27\%          & \textbf{44.39\%} & 94.07\%          \\
          Ours ($\delta=0.1$) & \textbf{89.56\%} & 30.46\%          & 98.13\%          \\
          \bottomrule
        \end{tabular}
      \end{sc}
    \end{small}
    \vskip -0.1in
  \end{table}

  \subsection{Experimental Results}

  We evaluate our method against baselines, including re-prompting \cite{Sun25Probing},
  linear steering \cite{bhalla2024towards} and hard replacement (using a Ground Truth
  probe).
  As shown in \cref{tab:method-comparison}, our method ($\delta=0.1$) achieves
  the highest accuracy ($89.56\%$). Crucially, these results serve as a causal
  validation of the IRST geometry rather than a mere performance boost. The ability
  to correct errors using only internal consistency confirms that the model retains
  correct latent information despite erroneous output quantization.

  Moreover, the sensitivity to $\delta$ corroborates the Noisy Quantization
  Model. While $\delta=0$ yields the highest correction rate (TP Corr: 44.39\%),
  it aggressively overwrites valid ambiguous states. The balanced performance at
  $\delta=0.1$ mirrors the theoretical ``bathtub'' stability, confirming that
  the representation space is indeed structured by a continuous potential
  subject to stochastic boundary noise. Full tolerance ablations and additional
  implementation details are deferred to Appendix~\ref{app:algo}.

  \begin{table}[t]
    \caption{Ablation study disentangling raw-sum and carry information for the
    dual-stream correction mechanism ($\delta=0$, Qwen3-4B, 3-term addition).}
    \label{tab:ablation_components}
    \centering
    \begin{minipage}{0.98\columnwidth}
      \centering
      \begin{small}
        \begin{sc}
          \begin{tabular}{lccc}
            \toprule Method & Token Acc & TP Corr & FP Pres \\
            \midrule R+C & 86.7\% & 42.3\% & 93.9\% \\
            R+TC         & 96.0\% & 69.3\% & 99.0\% \\
            TR+C         & 90.5\% & 65.4\% & 94.6\% \\
            \bottomrule
          \end{tabular}
        \end{sc}
      \end{small}
      \vspace{0.03in}

      \raggedright\footnotesize R+C denotes \textit{Raw-Sum
      Probe + Carry Probe}, R+TC denotes \textit{Raw-Sum Probe + True Carry},
      and TR+C denotes \textit{True Raw-Sum + Carry Probe}.
    \end{minipage}
    \vskip -0.1in
  \end{table}

  We further conduct an ablation study to decouple the geometric components used
  by the dual-stream protocol. As shown in \cref{tab:ablation_components},
  \textit{R+TC} reaches 96.0\% token accuracy, indicating that the model's latent
  local computation remains highly robust even when the final output token is
  wrong. In contrast, \textit{TR+C} only reaches 90.5\% token accuracy,
  mechanically isolating the noisy carry potential as the main bottleneck behind
  arithmetic failures.

  \section{Related Works}

  \subsection{LLM Arithmetic and Probing}
  While LLMs demonstrate impressive arithmetic capabilities \citep{Cobbe21GSM8K,Hendrycks21MATH,yuan2023well,zhang2024interpreting},
  the underlying mechanisms remain debated. One paradigm views arithmetic as discrete
  symbolic manipulation, where inputs are mapped to categorical sum types \citep{Quirke24Understanding}
  or rely on optimized tokenization schemes \citep{singh2024tokenization} to
  execute algorithmic logic. Conversely, heuristic analyses identify specific
  bottlenecks in this process, such as the narrow ``lookahead'' window for carry
  propagation \citep{Baeumel25Lookahead} and the fragility of multi-digit
  attention \citep{Qiu24Dissecting}. To bridge these views, recent probing studies
  have successfully mapped the layer-wise evolution of arithmetic signals \citep{yan2025addition,Zhu25Language}
  and utilized lightweight classifiers to detect error states or decode ground
  truth from residual streams \citep{Sun25Probing,hedstrom2025to}. However, these
  approaches predominantly treat internal states as discrete variables. Our work
  challenges this assumption by reinterpreting arithmetic generation as the
  noisy quantization of a \textit{Continuous Carry Potential}, providing a
  mechanistic explanation for why linear probes can successfully disentangle
  these signals.

  \subsection{Geometry of Reasoning}

  Broader studies on Riemannian geometry and topological persistence offer a foundational
  backdrop \citep{Shao18Riemannian,Naitzat20Topology,Fitz24Hidden,azizian2025geometries,tiblias2025shape},
  yet deciphering arithmetic requires analyzing specific algebraic structures.
  \citet{Nanda23Arithmetic} hypothesized that models perform modular arithmetic via
  rotational dynamics on circular manifolds. Recent refinements suggest LLMs
  utilize high-dimensional spirals or trigonometric encodings \citep{kantamneni2025language},
  often composing digit-wise circular representations with linear magnitude
  components \citep{levy2025language, engelsnot}. However, abstract topologies rarely
  map directly to concrete failure modes. Our framework refines these helical
  hypotheses by revealing a locally stratified manifold where the proximity of
  carry fibers determines error probabilities via mechanistic \textit{geometric
  slippages}.

  \section{Conclusion and Discussion}
  \label{sec:discussion}

  This work identifies the \textbf{IRST} as the geometric backbone of arithmetic,
  where errors manifest as mechanistic \textit{geometric slippages}. Our \textbf{Noisy Quantization Model} formalizes this process,
  showing that models often maintain a correct continuous \textit{Carry
  Potential} internally but fail during discrete token selection due to neural noise
  near quantization boundaries. Empirically, we observe that error distributions
  are predominantly off-by-one ($\pm 1$ shifts). Our geometric framework provides a structural explanation: unlike
  a circular clock-face representation, Anchors 0 and 9 are geometrically
  distant along the backbone. This helical geometry imposes a high energetic barrier
  against wrap-around and non-unit errors, effectively restricting failure modes
  to local shifts along a continuous IRST fiber.

 Our methodology leverages the single-digit tokenization of models (e.g., Qwen3
 \cite{yang2025qwen3}, Gemma3 \cite{team2025gemma}), which allows for a direct
 alignment between visual digits and internal arithmetic states without the
 interference of multi-digit grouping. While we hypothesize that BPE-based models
 rely on similar latent geometries, their carry signals are likely entangled
 within dense multi-digit embeddings, presenting a frontier for future research.
 At the same time, our current evidence still combines UMAP visualization and probe-based decoding rather than a full circuit-level
 localization of the arithmetic computation, and our empirical focus remains mainly on addition. Even so, the
 success of our dual-stream consistency check confirms that hallucinations are
 often quantization failures of a correct continuous potential, suggesting that
 improving LLM reliability depends on stabilizing the mapping from these
 continuous neural thoughts to discrete symbolic outputs.

\section*{Acknowledgement}
This work is supported in part by the National Natural Science Foundation of China (62576160), Fundamental Research Funds for the Central Universities (KG202508), 111 Center (B26023), and Fundamental and Interdisciplinary Disciplines Breakthrough Plan of the Ministry of Education of China (JYB2025XDXM118).

  \section*{Impact Statement}
  This paper presents work whose goal is to advance the field of Machine Learning.
  There are many potential societal consequences of our work, none which we feel
  must be specifically highlighted here.

  \bibliography{example_paper}
  \bibliographystyle{icml2026}

  %%%%%%%%%%%%%%%%%%%%%%%%%%%%%%%%%%%%%%%%%%%%%%%%%%%%%%%%%%%%%%%%%%%%%%%%%%%%%%%
  %%%%%%%%%%%%%%%%%%%%%%%%%%%%%%%%%%%%%%%%%%%%%%%%%%%%%%%%%%%%%%%%%%%%%%%%%%%%%%%
  % APPENDIX
  %%%%%%%%%%%%%%%%%%%%%%%%%%%%%%%%%%%%%%%%%%%%%%%%%%%%%%%%%%%%%%%%%%%%%%%%%%%%%%%
  %%%%%%%%%%%%%%%%%%%%%%%%%%%%%%%%%%%%%%%%%%%%%%%%%%%%%%%%%%%%%%%%%%%%%%%%%%%%%%%
  \newpage
  \appendix
  \onecolumn

  \section{Experimental Details for \cref{sec:represent}}
  \label{app:exp_details}

  \subsection{Model Configuration and Inference}
  We perform all experiments in the main text using the \textbf{Qwen3-4B} model \cite{yang2025qwen3}.
  All generations are performed using greedy decoding (temperature $T=0$) to
  eliminate stochasticity and ensure that the internal representations reflect
  the model's deterministic arithmetic reasoning path.

  \subsection{Dataset Construction}
  \label{subapp:dataset} Our dataset $\mathcal{D}$ consists of $N=10,000$
  distinct samples of 3-term addition problems ($A_{0}+ A_{1}+ A_{2}= S$), where
  each addend is a 10-digit integer.

  \textbf{Prompt Template.} To enforce deterministic arithmetic output and
  suppress conversational filler, we employ a \textit{forced-prefix} strategy. We
  apply the model's standard chat template to the instruction, and then
  explicitly append the arithmetic expression and an equals sign as the start of
  the assistant's response. An example from our dataset is shown below:

  \begin{tcolorbox}
    [colback=gray!10, colframe=gray!50, title=Prompt and Generation Example (Real
    Sample)] \small \texttt{[User Control Tokens]} \\
    \texttt{Calculate 651507672825 + 369229089834 + 294552136401. Only output a
    number. Don't output commas.} \\
    \\
    \texttt{[Assistant Control Tokens]} \\
    \texttt{651507672825 + 369229089834 + 294552136401 = } \textbf{\texttt{1315289000060}}
  \end{tcolorbox}
  \noindent
  In this example, the ground truth sum is \texttt{1315288899060}.

  \textbf{Truncated Generation Protocol.} To isolate the precise origin of
  arithmetic errors, we employ an \textit{early stopping} strategy. As shown in the
  example above, the model correctly generates the prefix \texttt{131528} but deviates
  at the next position (generating \texttt{9} instead of \texttt{8}). We
  terminate generation immediately at this first error. This ensures that every analyzed
  activation $\boldsymbol{h}_{p}^{(L)}$ is conditioned on a strictly correct history,
  confirming that the observed geometric slippage is a spontaneous divergence
  from the correct manifold rather than a downstream effect of error propagation.

  \textbf{Data Balancing.} Since random 10-digit addition yields imbalanced carry
  distributions, we employ stratified sampling to ensure the manifold at $p=4$
  is fully populated. We enforce a uniform distribution of $c_{p}\in \{0, 1, 2\}$
  at this target position to prevent visualization bias. Crucially, this intervention
  is strictly local; the natural carry distribution is preserved across all
  other positions, ensuring the dataset as a whole retains its global statistical
  representativeness.

  \subsection{Activation Extraction}
  \textbf{Tokenization Alignment.} We verify that the Qwen3 tokenizer treats
  each decimal digit as a single token in our prompt format. This one-to-one mapping
  between mathematical digits and semantic tokens is crucial for position-wise analysis.

  \textbf{Position Indexing.} Consistent with our mathematical framework, the generation
  index $p \in \{0, \dots, P-1\}$ proceeds from the most significant digit ($p=0$)
  to the least. Consequently, our primary analysis target $p=4$ corresponds to the
  5th generated token (mathematical significance $k=P-5$). We extract the
  residual stream activation $\boldsymbol{h}_{p}^{(L)}$ from the final layer
  $L=36$, taken after the final normalization layer and immediately prior to the
  unembedding projection.

  \subsection{Dimensionality Reduction Settings}
  \label{subapp:umap_params} We employ UMAP \cite{mcinnes2018umap} to project the
  2560-dimensional activation vectors into 2D. We utilize the \textit{cosine distance} metric to capture angular semantic relationships. To preserve the global
  topological structure while maintaining local separation, we set the hyperparameters
  to \texttt{n\_neighbors=300} and \texttt{min\_dist=0.3}. For visual clarity in
  the presented figures, we apply random downsampling to the projection to mitigate
  overplotting in high-density regions.

  \section{Comparison of Different Dimensionality Reductions}
  \label{app:dr_robustness}

  We performed a robustness check using Principal Component Analysis (PCA)
  \citep{abdi2010principal} and t-Distributed Stochastic Neighbor Embedding (t-SNE)
  \citep{maaten2008visualizing} as comparisons. We used the identical dataset of
  final-layer hidden states ($\boldsymbol{h}_{p}^{(L)}$ at position $p=4$) as presented
  in \cref{sec:represent}.

  \subsection{Quantitative Evaluation}

  We quantitatively assess the quality of the low-dimensional embeddings using
  the \texttt{trustworthiness} metric \citep{pedregosa2011scikit} with \texttt{n\_neighbors=50}.
  This metric evaluates how well the local neighborhood structure of the high-dimensional
  space is preserved in the 2D projection. As shown in Table \ref{tab:trustworthiness},
  both non-linear methods (UMAP and t-SNE) achieve near-perfect scores, significantly
  outperforming the linear PCA projection.

  \begin{table}[h]
    \centering
    \caption{Trustworthiness scores for different dimensionality reduction
    methods on $\boldsymbol{h}_{p}^{(L)}$ at $p=4$.}
    \label{tab:trustworthiness}
    \begin{tabular}{lc}
      \toprule \textbf{Method}              & \textbf{Trustworthiness} ($\uparrow$) \\
      \midrule PCA (Linear)                 & 0.8633                                \\
      t-SNE (Non-linear)                    & 0.9907                                \\
      \textbf{UMAP (Non-linear, Main Text)} & \textbf{0.9953}                       \\
      \bottomrule
    \end{tabular}
  \end{table}

  \subsection{Analysis of Visualizations}

\begin{figure*}[ht]
    \centering
    \begin{minipage}[b]{0.49\textwidth}
      \includegraphics[width=\linewidth]{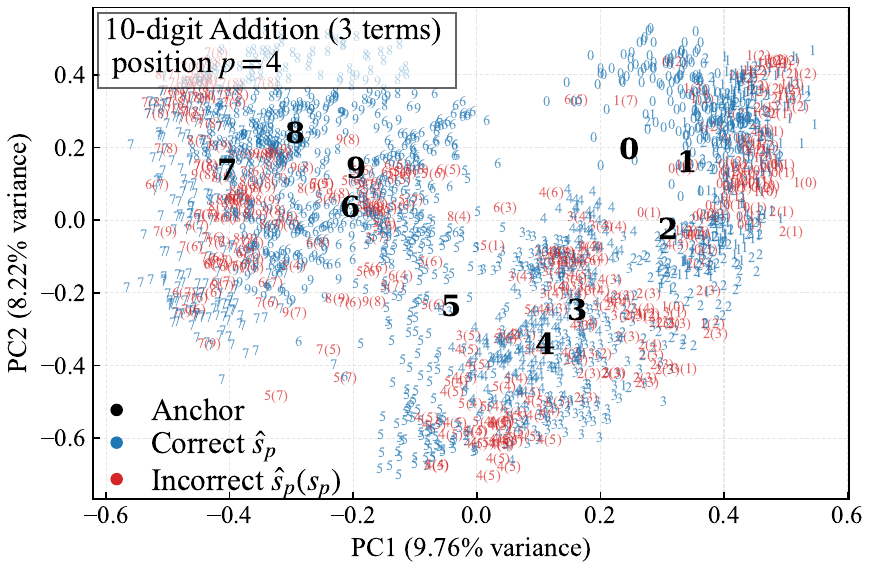}
    \end{minipage}
    \hfill
    \begin{minipage}[b]{0.49\textwidth}
      \includegraphics[width=\linewidth]{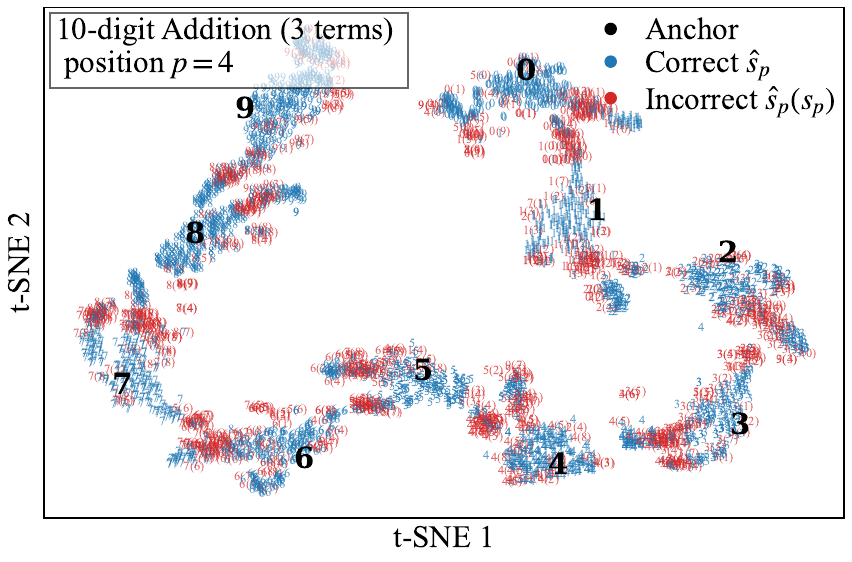}
    \end{minipage}
    \caption{\textbf{Alternative Dimensionality Reduction Visualizations.}.
    \textbf{(Left) PCA Projection:} Only macroscopic digit clustering is visible;
    the fine-grained IRST fibers are collapsed due to the low variance captured
    by the first two components (9.76\% and 8.22\%). \textbf{(Right) t-SNE
    Projection:} The visualization clearly recovers the same structural
    hierarchy (Macroscopic Backbone and Microscopic Texture) as UMAP, confirming
    that the IRST is not an artifact of a specific algorithm.}
    \label{fig:other_dimreduc}
  \end{figure*}

  Figure \ref{fig:other_dimreduc} presents the visualizations for PCA and t-SNE.

  \textbf{Linear vs. Non-linear:} The failure of PCA to reveal the backbone and texture
  structure is indicative of the non-linearity of the arithmetic manifold. The
  first two principal components explain only approximately 17.98\% of the total
  variance, suggesting that the relevant arithmetic information is distributed
  across a much higher-dimensional subspace. The significantly higher trustworthiness
  of non-linear methods confirms that local neighborhood preservation is essential
  for interpreting the model's internal states.

  \textbf{t-SNE vs. UMAP:} The t-SNE projection is qualitatively consistent with
  our UMAP results, exhibiting clear digit basins and stratified carry fibers.
  This cross-algorithm consistency provides strong evidence that the IRST is an
  intrinsic geometric feature of the residual stream. We chose UMAP for the main
  text because it marginally outperforms t-SNE in trustworthiness (0.9953 vs
  0.9907) and better preserves global continuity, which is critical for
  representing trajectories that span multiple digit basins.

  \section{Geometric Validation and Causal Steering}
  \label{app:geometric_validation}

The main text already presents a representative trajectory-level case study and
  causal steering result for $\mathcal{T}_{3}$ (\cref{fig:app_t3_analysis}). In
  this appendix, we restore the full trajectory-level interpretation of that
  figure before turning to the broader native-space ablations and
  intrinsic-dimensionality measurements.

  \subsection{Detailed Case Study on $\mathcal{T}_{3}$}

  We first visualize the latent geometry of this trajectory. We calculate the
  centroids of the stable clusters for $(3,3,0)$, $(4,4,1)$, and $(5,5,2)$ in
  the last-layer ($p=4$) representation space.

  \Cref{fig:app_t3_analysis} (Left) presents the projection of all samples
  belonging to $\mathcal{T}_{3}$. The x-axis represents the cosine distance
  relative to the central node centroid $(4,4,1)$, while the y-axis represents
  the analytical Carry Potential $\Phi$. The empirical distribution mirrors the
  theoretical schematic in the main text (\cref{fig:irst_analysis}): the
  low-potential cluster $(3,3,0)$, the central cluster $(4,4,1)$, and the
  high-potential cluster $(5,5,2)$ are connected by a clear V-shaped
  progression. Crucially, leakage errors like $(5,4,2)$ and hallucination
  errors like $(4,5,1)$ reside precisely in the transition zones between the
  stable basins. This confirms that these errors are geometric intermediate
  states where the representation has drifted away from the cluster center but
  has not yet crossed the decision boundary of the next carry fiber.

  \subsection{Detailed Causal Steering Dynamics}

  Using the same steering intervention defined in Section~\ref{sec:trajectory_validation},
  we can interpret \Cref{fig:app_t3_analysis} (Right) at the level of individual
  thresholds. The four representative samples are ordered exactly as their
  geometric positions would suggest:

  \begin{itemize}
    \item \textbf{Stable State $(5,5,2)$ ($\Phi=2.36$):} This sample is deeply
      embedded in the ``Carry 2'' basin. It requires a significant negative steering
      force ($\alpha \approx -0.5$) to suppress the carry signal and flip the output
      back to 4.

    \item \textbf{Hallucination Error $(4,5,1)$ ($\Phi=1.96$):} This sample
      represents a ``barely'' wrong 5. It lies extremely close to the decision
      boundary and requires only a minimal push ($\alpha \approx -0.1$) to
      correct it back to the ground truth 4.

    \item \textbf{Leakage Error $(5,4,2)$ ($\Phi=2.00$):} This sample represents
      a ``barely'' wrong 4. It resides just below the threshold and is corrected
      to 5 with a positive push of $\alpha \approx 0.3$.

    \item \textbf{Stable State $(4,4,1)$ ($\Phi=1.62$):} This sample is stable
      in the ``Carry 1'' basin. It requires a strong positive force ($\alpha \approx
      0.5$) to induce a transition to 5.
  \end{itemize}

  The precise ordering of these transition thresholds quantitatively confirms
  that the model's arithmetic decision is strictly determined by the projection
  of the representation onto the Carry Potential direction.

  \subsection{Native-Space Validation Across Trajectories}

  \begin{figure}[t]
    \centering
    \centerline{\includegraphics[width=1.0\columnwidth]{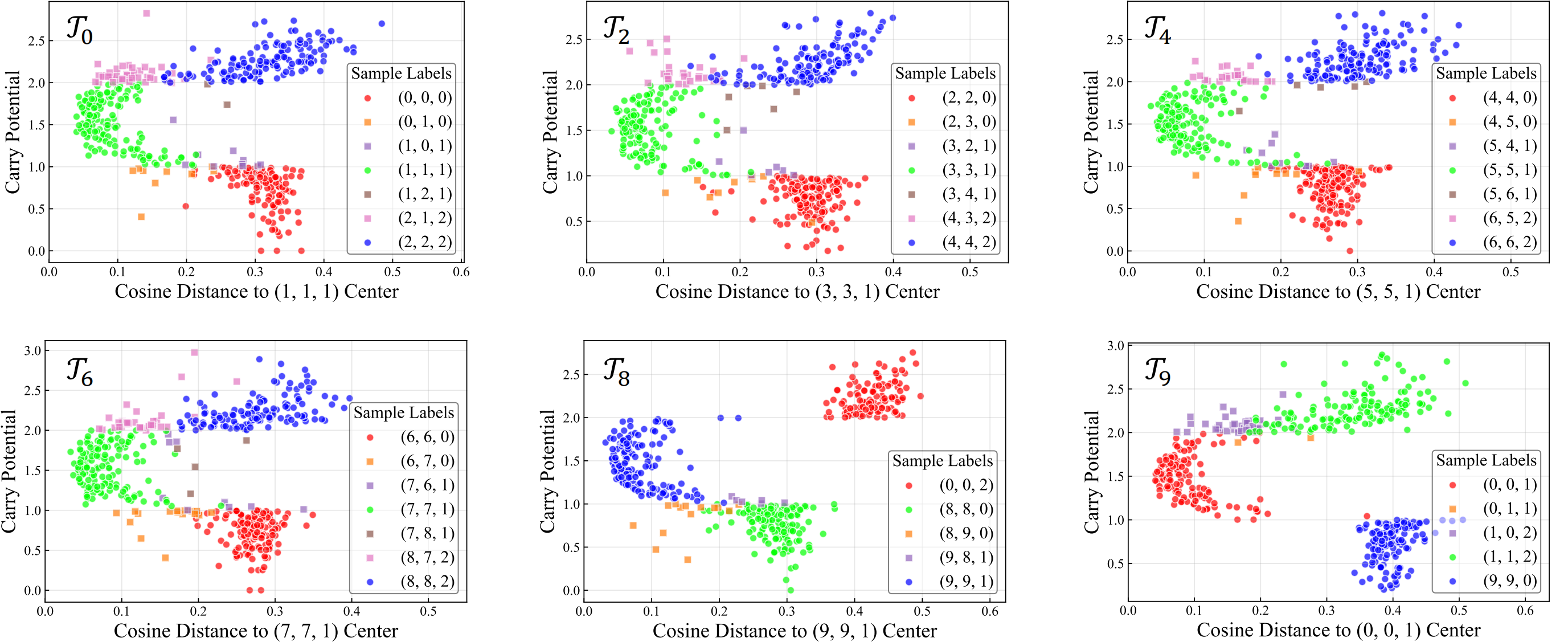}}
    \caption{\textbf{Projection-independent validation of the IRSTs in the native
    representation space.} Cosine distance to anchor states is evaluated directly
    in $\mathbb{R}^{2560}$ across multiple trajectories. The characteristic V-shaped
    correlation between native-space distance and the continuous carry potential
    $\Phi$ generalizes across $\mathcal{T}_{0}\dots \mathcal{T}_{9}$, supporting
    the claim that the IRST organization is not solely a 2D projection artifact.
    The exceptional behavior near $\mathcal{T}_{8}$ and $\mathcal{T}_{9}$ is
    consistent with the absence of a continuous adjacent-digit bridge across the
    $0 \leftrightarrow 9$ boundary.}
    \label{fig:vshape_native} \vskip -0.1in
  \end{figure}

  As an ablation on the projection choice, we also measure the trajectory geometry
  directly in the original residual space. \Cref{fig:vshape_native} shows that
  the native-space cosine-distance profiles preserve the same V-shaped correlation
  with $\Phi$ across $\mathcal{T}_{0}\dots\mathcal{T}_{9}$. In other words, the
  core ordering is already present in $\mathbb{R}^{2560}$ and does not depend on
  a particular 2D UMAP layout.

  \subsection{Intrinsic-Dimensionality Analysis}

  \begin{figure}[t]
    \centering
    \centerline{\includegraphics[width=0.50\columnwidth]{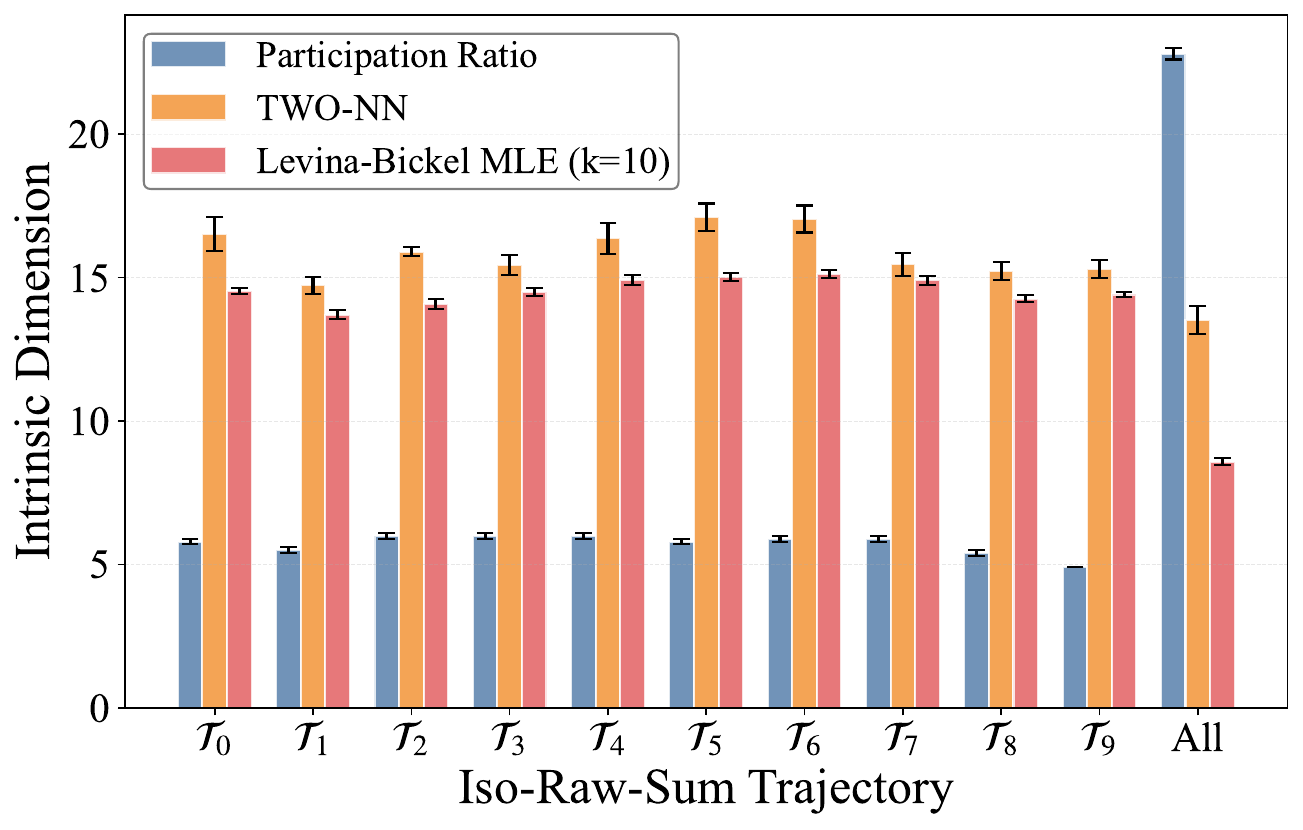}}
    \caption{\textbf{Intrinsic dimensionality across IRSTs.} We report the participation
    ratio, TWO-NN, and Levina--Bickel MLE estimates for $\mathcal{T}_{0}\dots\mathcal{T}_{9}$
    and their pooled union. The trajectory-conditioned subsets remain stably
    low-dimensional in native space, while the pooled set exhibits a larger linear
    effective dimension.}
    \label{fig:irst_id} \vskip -0.1in
  \end{figure}

  \Cref{fig:irst_id} summarizes intrinsic-dimensionality estimates computed
  directly in native space. Across $\mathcal{T}_{0}\dots\mathcal{T}_{9}$, the
  trajectory-conditioned subsets remain consistently low-dimensional, while the
  pooled set exhibits a noticeably larger linear effective dimension. This pattern
  supports the picture that individual IRSTs occupy relatively compact geometric
  subsets embedded within a broader shared arithmetic scaffold.

  \begin{figure}[t]
    \centering
    \centerline{\includegraphics[width=0.7\columnwidth]{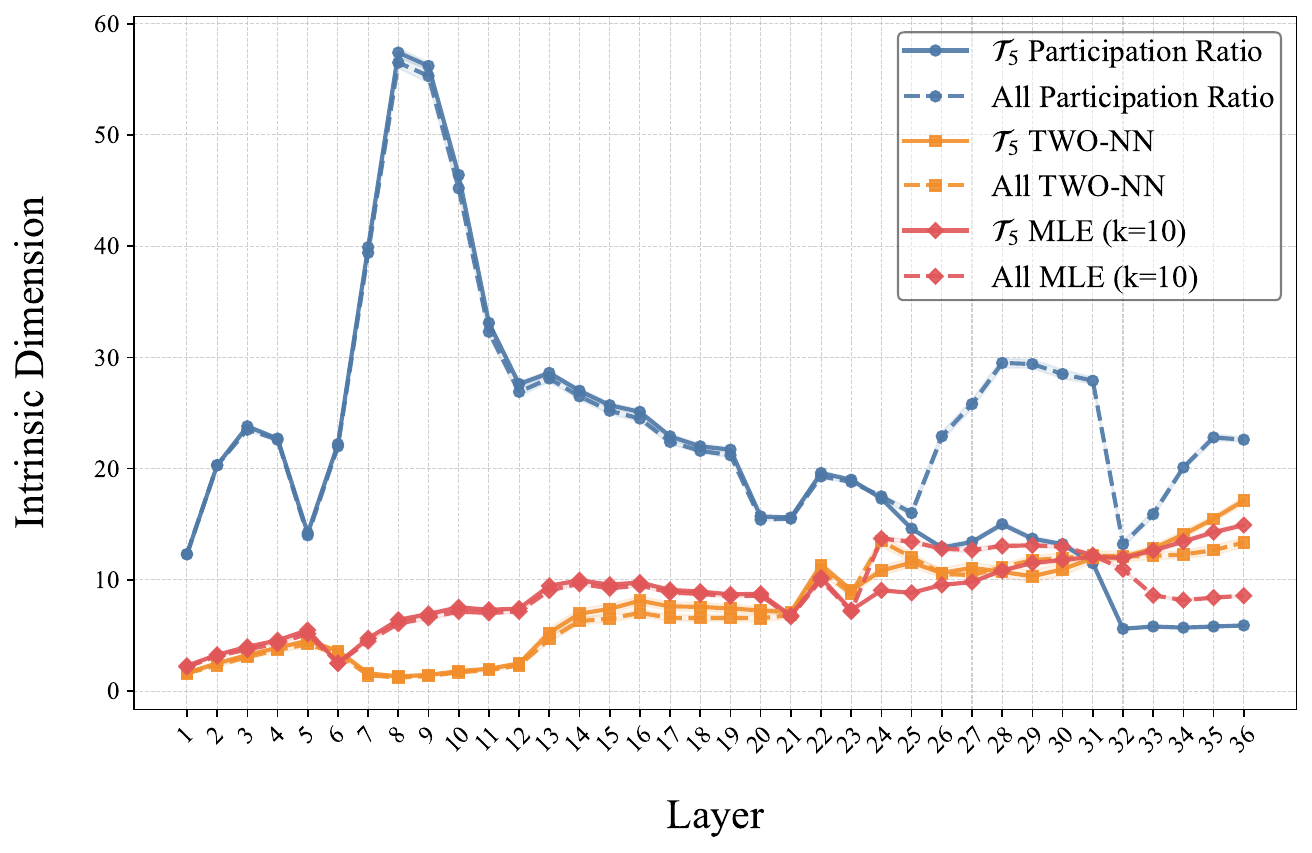}}
    \caption{\textbf{Layer-wise evolution of intrinsic dimensionality.} We compare
    the nonlinear and linear intrinsic-dimension estimates for $\mathcal{T}_{5}$
    and the pooled set of all trajectories across layers. The two remain broadly
    similar in early layers, while later layers develop more trajectory-specific
    local structure.}
    \label{fig:layer_id} \vskip -0.1in
  \end{figure}

  As further illustrated in \cref{fig:layer_id}, we additionally examined
  layer-wise intrinsic-dimension estimates for $\mathcal{T}_{5}$ and the pooled
  set (All). The two are broadly similar before layer 23. Between layers 23--31,
  $\mathcal{T}_{5}$ has slightly lower nonlinear ID (MLE) than All, while after
  layer 31 this trend reverses. One possible reading is that a relatively stable
  global scaffold forms first, followed by later refinement of
  trajectory-specific local structure, broadly consistent with Appendix~\ref{app:probe}.

  \section{Validation of the Raw Sum Assumption}
  \label{app:raw_sum_validation}

  In the main context, we posit that the model's arithmetic errors are primarily
  driven by latent carry deviations rather than failures in local column-wise
  summation. This assumption, mathematically expressed as
  $\hat{r}_{p}\approx r_{p}$, is foundational to both our theoretical derivation
  of latent carry states and the design of the inference-time correction mechanism.

  To validate this, we trained an 3-layer MLP classification probe to predict
  the absolute raw sum $r_{p}= \sum_{i}a_{i,p}$ from the final layer activations
  $\boldsymbol{h}_{p}^{(L)}$ at $p=4$. For our 3-term addition task, $r_{p}$
  ranges from 0 to 27, constituting a 28-class classification problem. Crucially,
  we evaluated this probe on a balanced dataset containing equal numbers of correct
  and incorrect model generations.

  \paragraph{Results.}
  The probe achieved an Accuracy of \textbf{89.35\%} and an AUC of \textbf{99.59\%}
  on the validation set, far exceeding the random baseline (Accuracy
  $\sim 3.6\%$). While this performance is not perfect—implying that a minority
  of errors may indeed stem from genuine local calculation failures—the high
  fidelity supports two critical conclusions.

  On the theoretical front, it indicates that the model predominantly encodes the
  correct local sum even when it generates an incorrect output digit $\hat{s}_{p}$.
  This suggests that arithmetic errors are largely geometrically constrained: the
  representation typically remains anchored to the correct \textbf{IRST} (preserving
  $r_{p}$) but slides along it into an incorrect carry fiber. Consequently,
  mapping output errors to carry deviations ($\hat{c}_{p}= \hat{s}_{p}- r_{p}$)
  is physically justified as the dominant error mechanism, though not an exclusive
  one. On the methodological side, the robustness of the internal raw sum signal
  validates the \textit{Local Calculation Stream} in our dual-stream
  intervention method (\cref{sec:method_correction}), ensuring that extracting $\hat
  {r}_{p}$ serves as a dependable, albeit probabilistic, anchor for reconstruction
  in the vast majority of instances.

  \begin{figure}[t]
    \centering
    \centerline{\includegraphics[width=0.5\columnwidth]{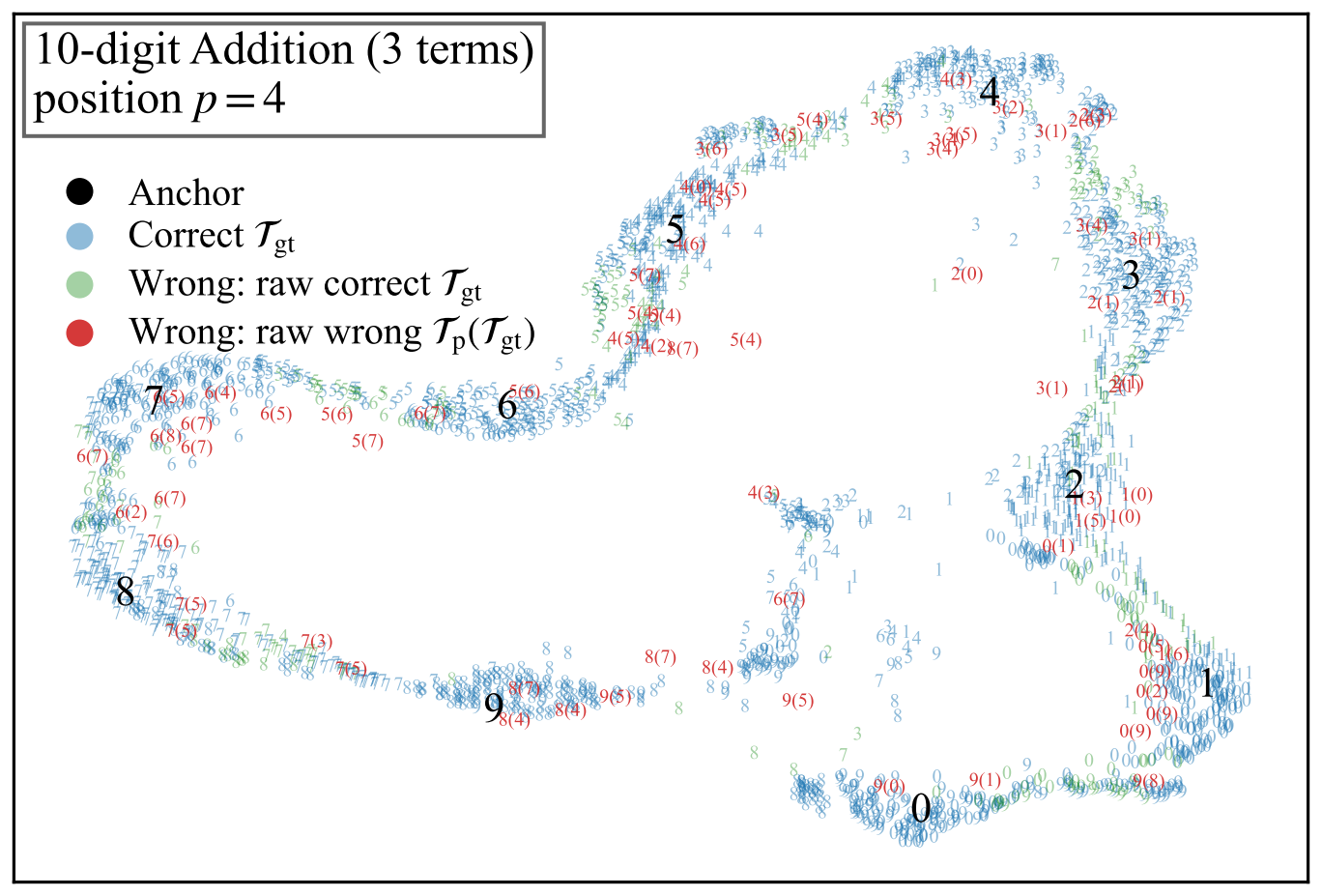}}
    \caption{\textbf{Geometric signatures of carry-based and non-carry errors.}
    Error modes are decoupled using the raw-sum probe. Carry-based errors, where
    the local raw sum is still recovered correctly, remain concentrated near adjacent
    decision boundaries. In contrast, non-carry errors scatter across more distant
    regions and do not follow a single continuous trajectory, indicating a distinct
    failure mode beyond the dominant carry-deviation mechanism. In the plotted
    annotations, $\mathcal{T}$ denotes the inferred trajectory identity, $gt$
    denotes the ground-truth digit, and $p$ denotes the analyzed generation
    position (fixed here at $p=4$).}
    \label{fig:rawsum_error_modes} \vskip -0.1in
  \end{figure}

  This distinction is visualized directly in \cref{fig:rawsum_error_modes}. The
  carry-based subset follows the same boundary-concentrated pattern emphasized in
  the main text, whereas the raw-sum-failure subset is geometrically far less structured.

  \section{Derivation of Latent Carry States}
  \label{app:latent_carry}

  In this section, we provide the detailed derivation for inferring the latent predicted
  input carry $\hat{c}_{p}$ from the observed model output $\hat{s}_{p}$. This
  inference is grounded in the modular arithmetic properties of the addition
  task.

  \paragraph{Problem Setup.}
  Recall the modular relationship for the ground truth digit $s_{p}$, derived
  from the raw sum $r_{p}$ and the ground truth input carry $c_{p}$:
  \begin{equation}
    \label{eq:gt_identity}s_{p}\equiv (r_{p}+ c_{p}) \pmod{10}.
  \end{equation}
  We posit that the model generates its predicted digit $\hat{s}_{p}$ using the same
  logic but based on its internal states. Assuming the local summation is robust
  (i.e., the perceived raw sum matches the true raw sum $\hat{r}_{p}= r_{p}$), any
  deviation in the output must stem from a latent predicted carry $\hat{c}_{p}$:
  \begin{equation}
    \label{eq:pred_identity}\hat{s}_{p}\equiv (r_{p}+ \hat{c}_{p}) \pmod{10}.
  \end{equation}
  We analyze the carry implications for the two primary error modes: Underestimation
  and Overestimation.

  \paragraph{1. Underestimation (Leakage).}
  This error mode occurs when the model outputs a digit that is exactly 1 less
  than the ground truth (modulo 10). Mathematically, this is expressed as:
  \begin{equation}
    \hat{s}_{p}\equiv (s_{p}- 1) \pmod{10}.
  \end{equation}
  Substituting the definitions from \cref{eq:gt_identity} and \cref{eq:pred_identity}
  into this relation:
  \begin{align}
    (r_{p}+ \hat{c}_{p}) & \equiv (r_{p}+ c_{p}- 1) \pmod{10}\nonumber \\
    \hat{c}_{p}          & \equiv (c_{p}- 1) \pmod{10}.
  \end{align}
  Since carry values in standard addition are small integers (typically $c_{p}\in
  \{0, 1, 2\}$ for 3 terms addition), this equivalence implies $\hat{c}_{p}= c_{p}
  - 1$. Thus, an output underestimation directly maps to a \textit{Leakage} of
  the internal carry.

  \paragraph{2. Overestimation (Hallucination).}
  This error mode occurs when the model outputs a digit that is exactly 1 greater
  than the ground truth (modulo 10):
  \begin{equation}
    \hat{s}_{p}\equiv (s_{p}+ 1) \pmod{10}.
  \end{equation}
  Substituting the definitions yields:
  \begin{align}
    (r_{p}+ \hat{c}_{p}) & \equiv (r_{p}+ c_{p}+ 1) \pmod{10}\nonumber \\
    \hat{c}_{p}          & \equiv (c_{p}+ 1) \pmod{10}.
  \end{align}
  This implies $\hat{c}_{p}= c_{p}+ 1$. Thus, an output overestimation maps to a
  \textit{Hallucination} of an additional carry unit.

  % ---------------------------------------------------------

  \section{Derivation of the Noisy Quantization Error Rate}
  \label{app:error_derivation}

  We derive the error probability for the \textbf{Noisy Quantization Model}. Let
  $\Phi$ be the ground truth potential and $\hat{\Phi}= \Phi + \epsilon$ be the
  noisy estimate with $\epsilon \sim \mathcal{N}(0, \sigma^{2})$. The discrete carry
  is $\hat{c}= \lfloor \hat{\Phi}\rfloor$. We define the fractional part as
  $\delta(\Phi) = \Phi - \lfloor \Phi \rfloor$. The standard Q-function is defined
  as:
  \begin{equation}
    Q(z) = \frac{1}{\sqrt{2\pi}}\int_{z}^{\infty}e^{-t^2/2}\, dt.
  \end{equation}

  \paragraph{1. Underestimation (Leakage).}
  Leakage occurs when the noise pushes the potential below the integer floor, i.e.,
  $\Phi + \epsilon < \lfloor \Phi \rfloor$, which simplifies to
  $\epsilon < -\delta(\Phi)$. Utilizing the symmetry of the Gaussian distribution
  ($P(Z < -z) = Q(z)$), the probability is:
  \begin{equation}
    P_{\text{leak}}(\Phi) = P\left(\frac{\epsilon}{\sigma}< -\frac{\delta(\Phi)}{\sigma}
    \right) = Q\left(\frac{\delta(\Phi)}{\sigma}\right).
  \end{equation}

  \paragraph{2. Overestimation (Hallucination).}
  Hallucination occurs when the noise pushes the potential above the next integer
  ceiling, i.e., $\Phi + \epsilon \geq \lfloor \Phi \rfloor + 1$, which implies $\epsilon
  \geq 1 - \delta(\Phi)$. The probability is directly given by the Q-function:
  \begin{equation}
    P_{\text{halluc}}(\Phi) = P\left(\frac{\epsilon}{\sigma}\geq \frac{1 -
    \delta(\Phi)}{\sigma}\right) = Q\left(\frac{1 - \delta(\Phi)}{\sigma}\right).
  \end{equation}

  \paragraph{Total Error Probability.}
  Assuming negligible probabilities for non-unit errors (valid for small
  $\sigma$), the total error rate is the sum of the leakage and hallucination
  risks:
  \begin{equation}
    P(\text{Error}\mid \Phi) = Q\left(\frac{\delta(\Phi)}{\sigma}\right) + Q\left
    (\frac{1 - \delta(\Phi)}{\sigma}\right).
  \end{equation}

  \begin{figure*}[!t]
    % \vskip 0.2in
    \begin{center}
      \begin{minipage}[t]{0.48\textwidth} % 单图占0.48倍列宽（留间距）
        \centering
        \includegraphics[width=0.9\linewidth]{
          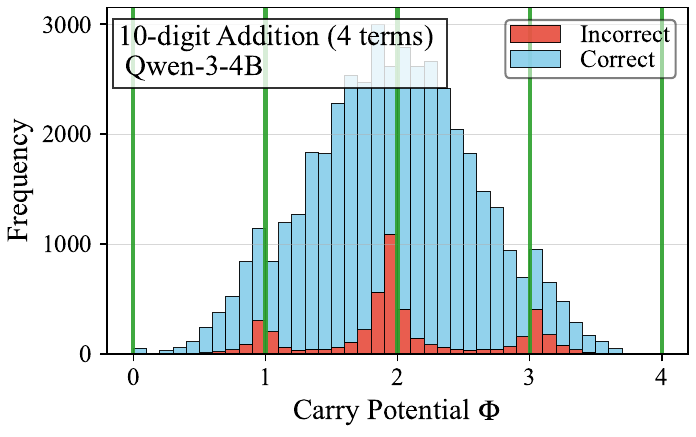
        }
      \end{minipage}
      \hfill % 第一排左右图间距
      \begin{minipage}[t]{0.48\textwidth}
        \centering
        \includegraphics[width=0.9\linewidth]{
          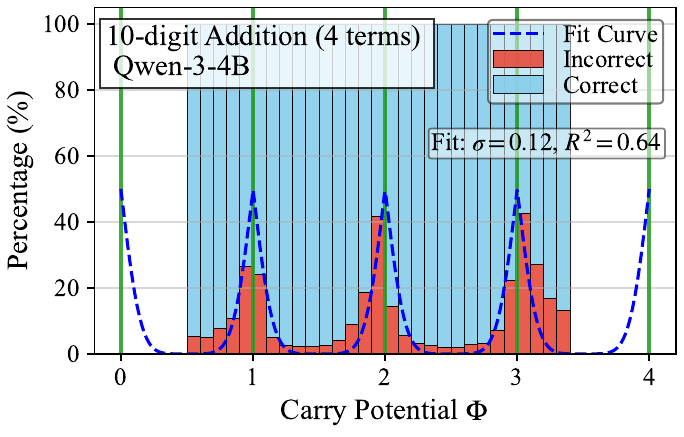
        }
      \end{minipage}
      \vspace{0.3cm}
      \begin{minipage}[t]{0.48\textwidth}
        \centering
        \includegraphics[width=0.9\linewidth]{
          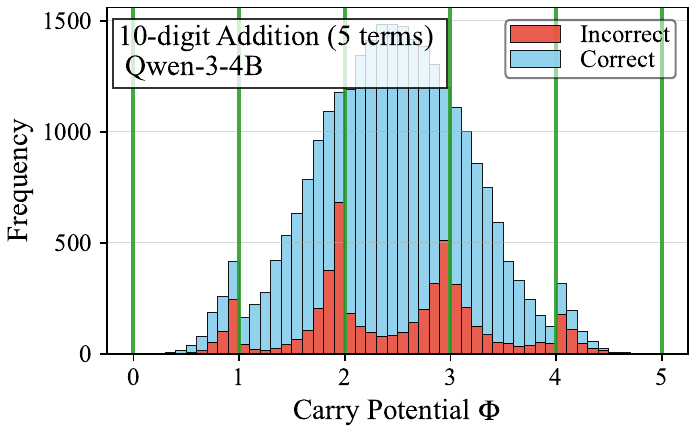
        }
      \end{minipage}
      \hfill % 第二排左右图间距
      \begin{minipage}[t]{0.48\textwidth}
        \centering
        \includegraphics[width=0.9\linewidth]{
          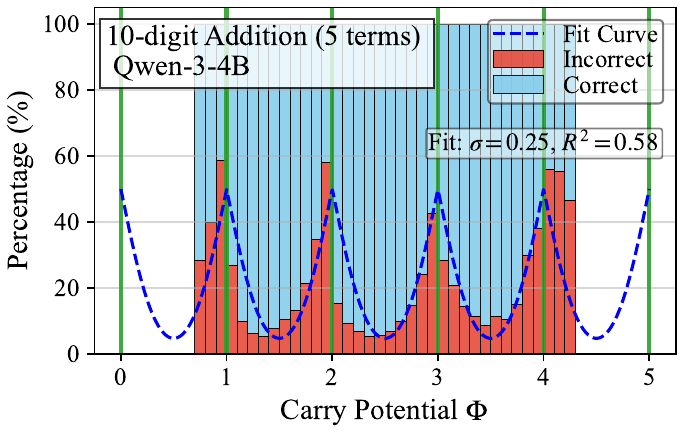
        }
      \end{minipage}
      \caption{ \textbf{Scaling of cognitive noise with task complexity.} We
      extend the analysis of \cref{fig:buthtub1} to 4-term ($n=4$) and 5-term ($n
      =5$) addition. The periodic bathtub structure persists across settings,
      validating the universality of the quantization mechanism. However, the
      fitted noise level $\sigma$ increases significantly from 0.05 ($n=3$) to
      0.12 ($n=4$) and 0.25 ($n=5$), indicating a sharp degradation in signal fidelity
      as the operand count grows. }
      \label{fig:buthtub2}
    \end{center}
  \end{figure*}

  \section{Scaling Dynamics of Cognitive Noise}
  \label{app:noise_scaling}

  We extend our analysis to higher-complexity tasks to investigate how the cognitive
  noise $\sigma$ scales. As shown in \cref{fig:buthtub2}, the periodic \textit{bathtub}
  error distribution is preserved in both 4-term and 5-term addition, confirming
  that the quantization of a continuous carry potential is a universal mechanistic
  primitive in LLM arithmetic. Notably, we observe a sharp non-linear growth in $\sigma$
  as the number of addends ($n$) increases: from $\sigma \approx 0.05$ at $n=3$
  to $\sigma \approx 0.25$ at $n=5$. This increase in noise causes the overlap of
  Gaussian tails near the quantization thresholds, effectively eliminating the
  \textit{robust plateaus} where error rates were previously near zero.

\begin{figure*}[t]
    \centering
    \begin{minipage}[t]{0.48\textwidth}
      \centering
      \includegraphics[width=0.9\linewidth]{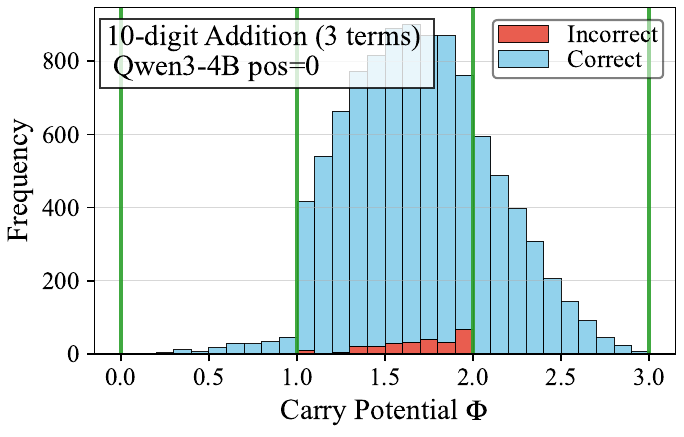}
    \end{minipage}
    \hfill
    \begin{minipage}[t]{0.48\textwidth}
      \centering
      \includegraphics[width=0.9\linewidth]{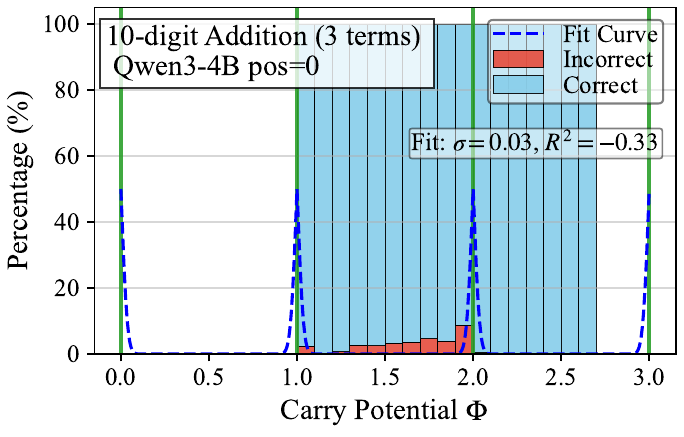}
    \end{minipage}
    \vspace{0.2cm}
    \begin{minipage}[t]{0.48\textwidth}
      \centering
      \includegraphics[width=0.9\linewidth]{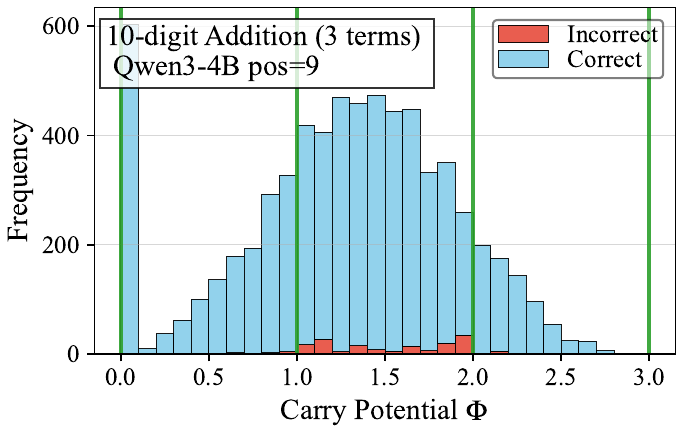}
    \end{minipage}
    \hfill
    \begin{minipage}[t]{0.48\textwidth}
      \centering
      \includegraphics[width=0.9\linewidth]{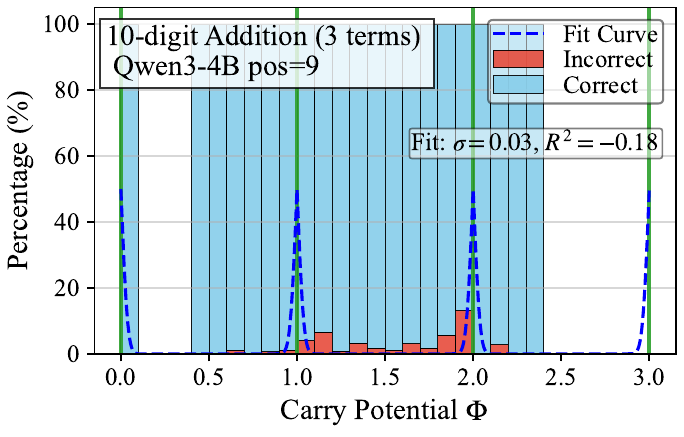}
    \end{minipage}
    \caption{\textbf{Boundary-position bathtub profiles.} Frequency and conditional-error
    distributions at the most significant digit ($p=0$, top row) and the least
    significant digit ($p=9$, bottom row) for 3-term 10-digit addition. Unlike
    the interior columns summarized in the main text, these boundary positions
    show structurally different profiles, highlighting that the steady-state bathtub
    regime is primarily an interior-position phenomenon.}
    \label{fig:boundary_bathtub}
  \end{figure*}

  Our analysis identifies three primary drivers of this noise. First, the \textit{operand
  count ($n$)} increases the informational load on the attention mechanism, diluting
  the signal-to-noise ratio during the aggregation of carry potential. Second, longer
  \textit{digit lengths ($m$)} incur a cumulative cognitive load, leading to higher
  baseline noise. Finally, while the results in the main text aggregate all positions
  to demonstrate the general mechanism, a boundary-focused analysis reveals that
  the edge positions ($p=0$ and $p=9$) deviate from the interior periodic regime.
  As shown in \cref{fig:boundary_bathtub}, these boundary columns exhibit distinct
  frequency and error profiles, consistent with reduced effective carry ambiguity
  at the most significant digit and stricter discrete constraints at the least
  significant digit. These findings suggest that the arithmetic capacity of Transformers
  is fundamentally bounded by the accumulation of this analog neural noise, a
  scaling property we leave for future systematic formulation.

  \section{Generalize to Other LLMs}
  \label{sec:other-llms}

  To validate the universality of our geometric framework, we extend our
  analysis beyond the Qwen3-4B model used in the main text. We visualize the last-layer
  representations of two additional models with different scales and architectures:
  \textbf{Qwen3-8B} and \textbf{Gemma-3-4B-IT}.

  \begin{figure*}[!t]
    \centering
    \begin{minipage}[b]{0.49\textwidth}
      \includegraphics[width=\linewidth]{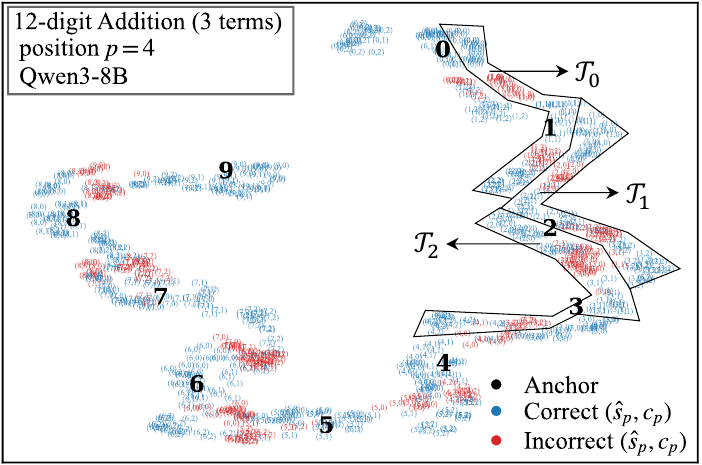}
    \end{minipage}
    \hfill
    \begin{minipage}[b]{0.49\textwidth}
      \includegraphics[width=\linewidth]{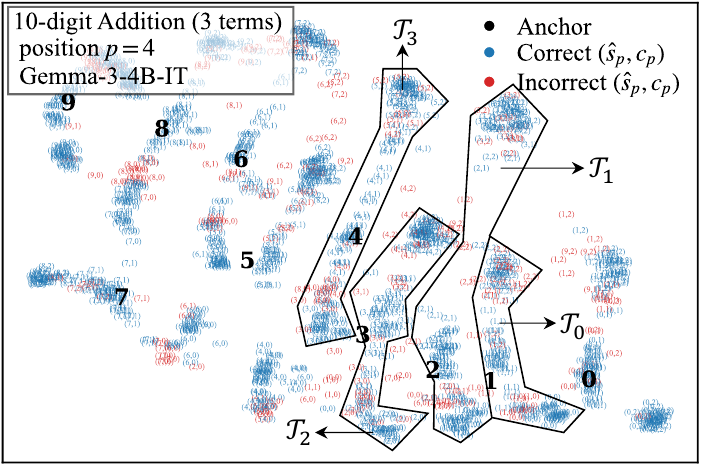}
    \end{minipage}
    \caption{ \textbf{Generalization of the IRST geometry across different
    models.} \textbf{(Left)} UMAP visualization for \textbf{Qwen3-8B} on a 12-digit
    addition task. The manifold structure is highly consistent with the 4B model,
    featuring a sequential arrangement of digit basins (0--9) connected by clear
    trajectories (e.g., $\mathcal{T}_{0}, \mathcal{T}_{1}, \mathcal{T}_{2}$).
    \textbf{(Right)} UMAP visualization for \textbf{Gemma-3-4B-IT} on a 10-digit
    addition task. Despite architectural differences, the fundamental geometry
    of IRSTs remains (e.g., $\mathcal{T}_{0}\sim \mathcal{T}_{3}$). }
    \label{fig:irst-other-llms}
  \end{figure*}

  \subsection{IRSTs}
  As shown in \cref{fig:irst-other-llms} (Left), the Qwen3-8B model exhibits a
  manifold geometry strikingly similar to its smaller counterpart Qwen3-4B in
  \cref{fig:umap1}. The digit anchors are arranged in a distinct, quasi-linear
  sequence from 0 to 9. The IRSTs form continuous paths that seamlessly bridge
  adjacent digit basins. This indicates that the geometric encoding strategy is stable
  across model scales within the same architectural lineage.

  The Gemma-3-4B-IT model (\cref{fig:irst-other-llms}, Right), representing a different
  model family, presents a slightly different global curvature. However, the
  core geometric properties predicted by our framework remain intact. We can clearly
  identify the basin structure for digits 0--9 and trace specific IRSTs.

  This picture is further supported by the \textbf{top} panel of
  \cref{fig:quanta_models}, which visualizes the single-task arithmetic
  transformers of \citet{quirke2024arithmetic}. The under-converged checkpoint
  still exhibits continuous inter-basin bridges similar to the IRST geometry in
  our main setting, whereas the fully converged checkpoint resolves into more
  isolated basins. This suggests that the same continuous organization can arise
  as an intermediate representational regime even in specialized arithmetic
  models.

  These observations suggest that our IRST is not an artifact of a specific
  model checkpoint but likely a convergent mechanism for arithmetic
  representation in state-of-the-art LLMs. Regardless of the specific architecture,
  models appear to learn to organize arithmetic states by ``threading'' raw-sum
  manifolds through semantic digit anchors.

  \begin{figure*}[!t]
    \centering
    \begin{minipage}[t]{0.48\textwidth}
      \centering
      \includegraphics[width=0.9\linewidth]{
        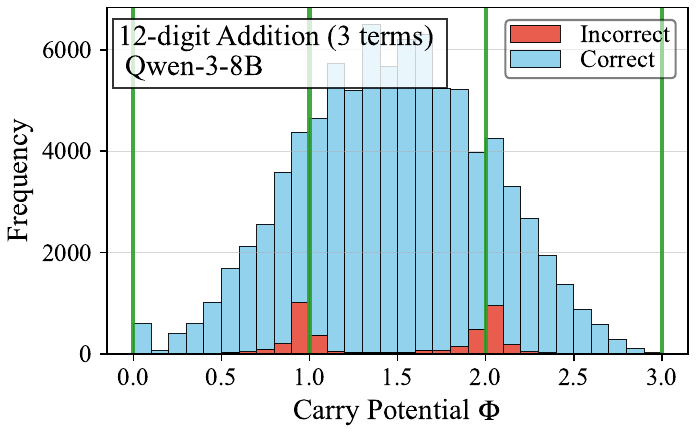
      }
    \end{minipage}
    \hfill
    \begin{minipage}[t]{0.48\textwidth}
      \centering
      \includegraphics[width=0.9\linewidth]{
        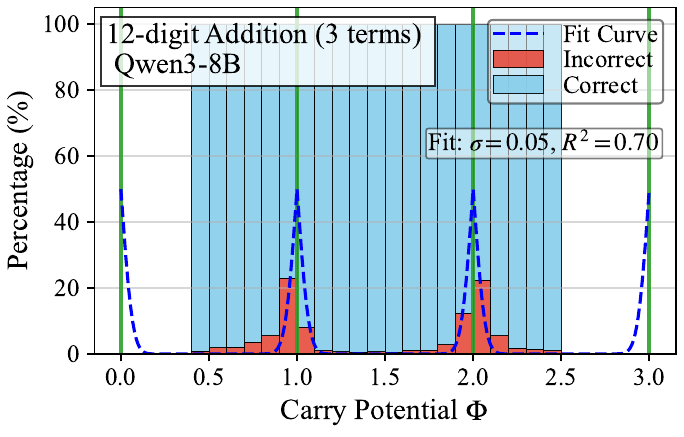
      }
    \end{minipage}
    \vspace{0.3cm} % 上下两排的间距
    \begin{minipage}[t]{0.48\textwidth}
      \centering
      \includegraphics[width=0.9\linewidth]{
        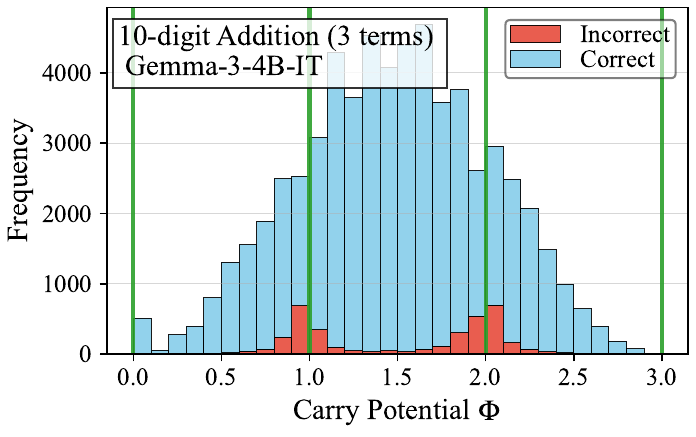
      }
    \end{minipage}
    \hfill
    \begin{minipage}[t]{0.48\textwidth}
      \centering
      \includegraphics[width=0.9\linewidth]{
        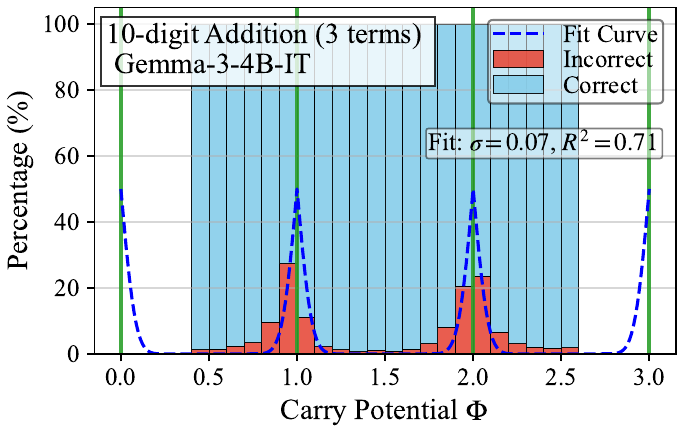
      }
    \end{minipage}

    \caption{ \textbf{Generalization of the Noisy Quantization hypothesis.}
    Validation on Qwen3-8B (12-digit addition, \textbf{top}) and Gemma-3-4B-IT (10-digit
    addition, \textbf{bottom}). The \textbf{Left} panels show the sample
    frequency distribution relative to Carry Potential $\Phi$, indicating that the
    dataset covers the entire potential space. The \textbf{Right} panels display
    the conditional error rates (red bars) overlaid with our theoretical fit (dashed
    blue line). The recurrence of the periodic error pattern near integer
    boundaries (green lines) and the high fitting scores ($R^{2}\approx 0.7$)
    confirm that the geometric quantization mechanism is robust across different
    model sizes and families. }
    \label{fig:buthtub-other-llms}
  \end{figure*}

  \subsection{Noisy Quantization Dynamics}
\begin{figure*}[t]
    \centering
    \begin{minipage}[b]{0.72\textwidth}
      \centering
      \includegraphics[width=\linewidth]{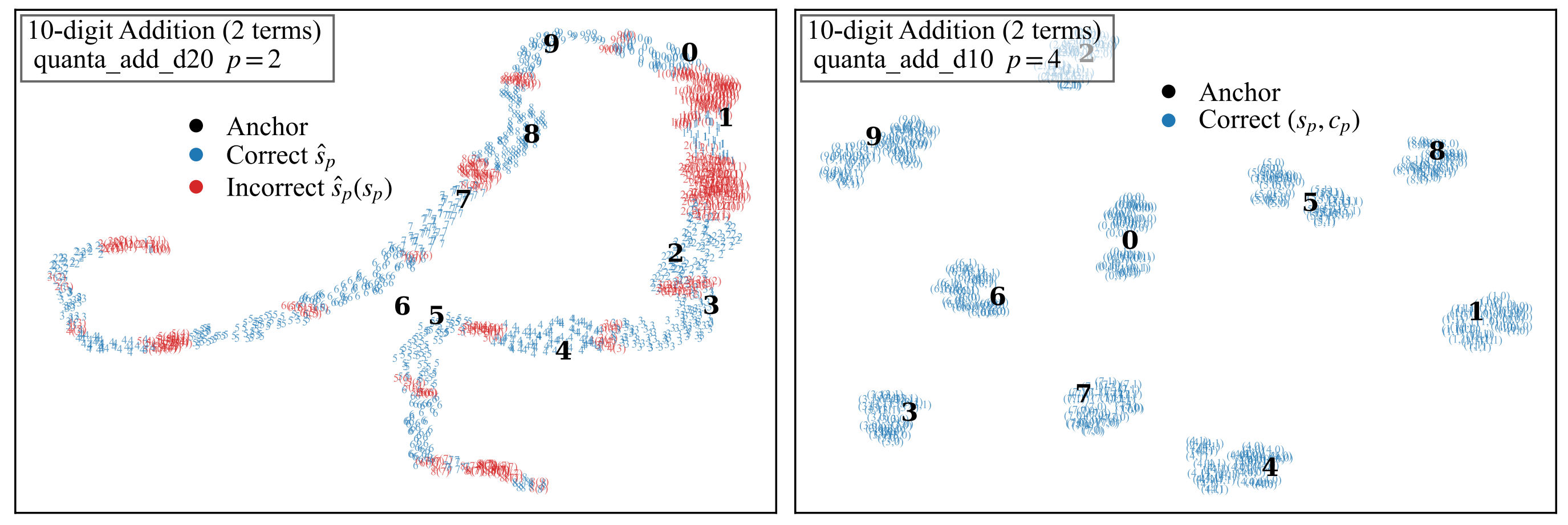}
    \end{minipage}
    \vspace{0.2cm}
    \begin{minipage}[b]{0.72\textwidth}
      \centering
      \includegraphics[width=\linewidth]{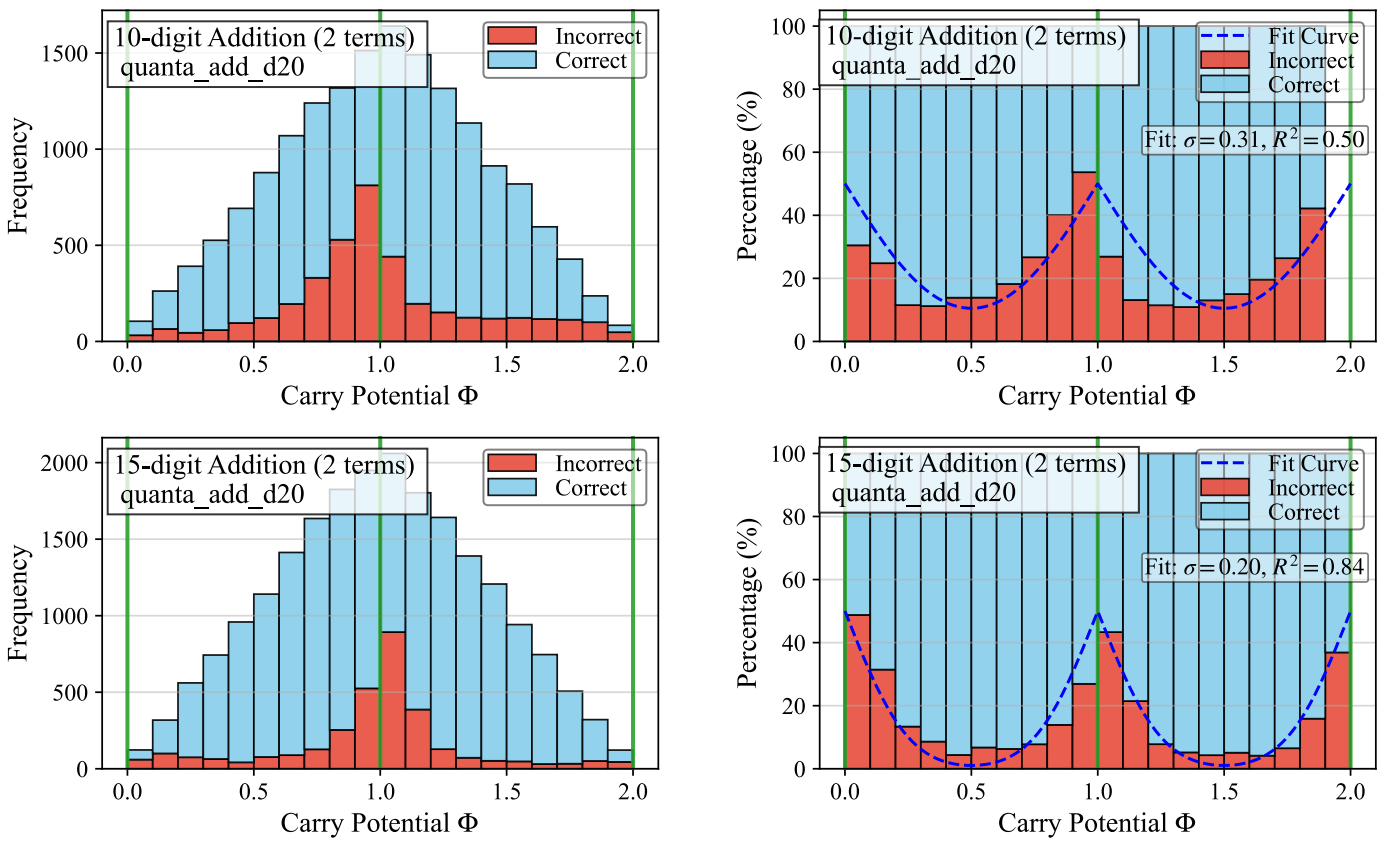}
    \end{minipage}
    \caption{\textbf{Additional evidence from specialized arithmetic transformers.}
    \textbf{(Top)} UMAP projections from the under-converged and fully converged
    single-task addition models reported by \citet{quirke2024arithmetic}. The
    under-converged model exhibits continuous inter-basin geometry similar to our
    main setting, while the fully converged model forms more isolated basins.
    \textbf{(Bottom)} Conditional error-rate curves for the under-converged model
    still follow the periodic bathtub pattern predicted by the Noisy Quantization
    Model.}
    \label{fig:quanta_models} \vskip -0.1in
  \end{figure*}

  We further validate the dynamic mechanism of error generation—the Noisy Quantization
  Model—across different model families. Figure \ref{fig:buthtub-other-llms}
  visualizes the relationship between the continuous Carry Potential $\Phi$ and
  the generation accuracy for Qwen3-8B and Gemma-3-4B-IT, while
  \cref{fig:quanta_models} extends the comparison to specialized arithmetic
  transformers trained solely on addition. Together, these results show that
  the same quantization-driven error landscape appears both in general-purpose
  LLMs and in task-specific arithmetic models before full convergence.

  \paragraph{Recurrence of the Bathtub Curve.}
  As shown in the right panels of Figure \ref{fig:buthtub-other-llms}, both Qwen-3-8B
  and Gemma-3-4B-IT exhibit the characteristic periodic error distribution
  observed in the main text. Error rates remain negligible near stable midpoints
  (e.g., $0.5, 1.5$) but spike sharply as $\Phi$ approaches integer boundaries ($1
  .0, 2.0$). This confirms that arithmetic failures are structurally induced by
  signal ambiguity near quantization thresholds rather than random stochasticity.
  The \textbf{bottom} panel of \cref{fig:quanta_models} shows the same
  bathtub-shaped error profile in the under-converged specialized arithmetic
  transformer, indicating that geometric slippage is not unique to
  general-purpose LLMs.

  \paragraph{Theoretical Fit and Noise Levels.}
  Fitting our theoretical error function (\cref{eq:double_q}) to the empirical data
  yields a robust fit ($R^{2}\geq 0.70$) for both models. We estimate the
  cognitive noise levels to be $\sigma \approx 0.05$ for Qwen-3-8B and
  $\sigma \approx 0.07$ for Gemma-3-4B-IT. Notably, Qwen-3-8B is evaluated on a
  more challenging 12-digit addition task compared to the 10-digit task used for
  the 4B models. The fact that Qwen-3-8B maintains a similarly low noise level ($\sigma
  \approx 0.05$) despite the increased sequence length indicates superior arithmetic
  precision, as the smaller model would likely exhibit significantly higher
  noise if subjected to this harder regime. In the same spirit, the specialized
  arithmetic transformers in \cref{fig:quanta_models} provide a qualitative
  training-time contrast: the under-converged checkpoint still exhibits the
  bathtub-shaped failure profile, whereas the fully converged checkpoint shows
  more isolated basins in representation space, suggesting that full convergence
  sharpens the quantization geometry and suppresses boundary ambiguity.
  Ultimately, the universal emergence
  of the \textit{bathtub} curve suggests that the Noisy Quantization Model is a fundamental
  phenomenological account of arithmetic processing in LLMs.

  \begin{figure*}[ht]
    \centering
    \begin{minipage}[b]{0.55\textwidth}
      \includegraphics[width=\linewidth]{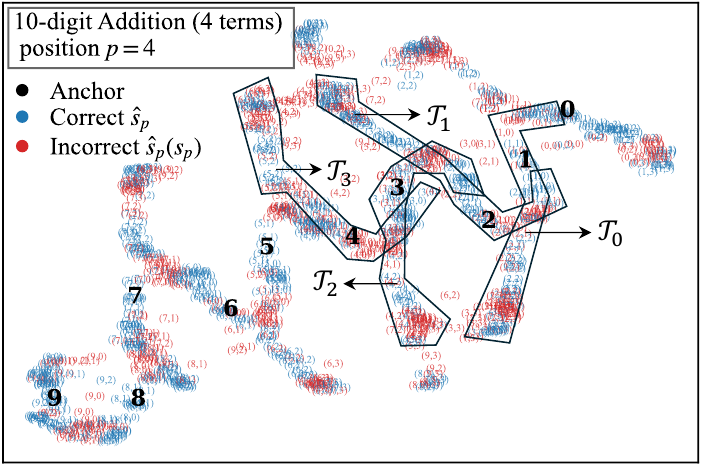}
    \end{minipage}
    \hfill % Automatically fill space
    \begin{minipage}[b]{0.35\textwidth}
      \includegraphics[width=\linewidth]{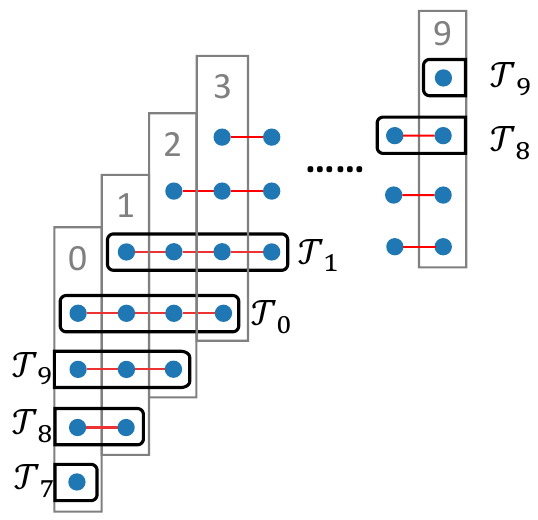}
    \end{minipage}
    \caption{ \textbf{IRST analysis for 4-term addition ($c_{p}\in \{0, 1, 2, 3\}$).}
    \textbf{(Left)} 2D UMAP projection of activations at $p=4$. While the digit
    backbone (0-9) persists, the increased density of carry fibers causes visual
    entanglement of trajectories in 2D space. However 3D plots can resolve these
    overlaps. \textbf{(Right)} Schematic of the expanded geometry. Between any adjacent
    digit basins (e.g., 0 and 1), there are now three distinct transition paths
    (red lines) corresponding to different raw sums, representing the increased risk
    of carry confusion. }
    \label{fig:irst-num4} \vskip -0.1in
  \end{figure*}

  \section{Generalize to 4-Term Addition}
  \label{app:four_term_addition}

  In the main text, we focused on 3-term addition where the input carry $c_{p}$
  is limited to $\{0, 1, 2\}$. To demonstrate the universality of the IRST framework,
  we extend our analysis to the addition of four 10-digit integers. In this setting,
  the maximum possible carry is 3, meaning the internal state must encode four distinct
  carry fibers $c_{p}\in \{0, 1, 2, 3\}$ within each digit basin.

  \subsection{Visualizing Higher-Dimensional Manifolds}

  Figure \ref{fig:irst-num4} (Left) presents the UMAP projection of the final
  layer activations ($p=4$) for the 4-term task. While the macroscopic backbone which
  is defined by the sequence of digit anchors $0 \to 9$ remains intact, the
  microscopic texture becomes significantly denser compared to the 3-term case.

  We observe that distinct IRSTs (e.g., $\mathcal{T}_{0}, \mathcal{T}_{1}, \mathcal{T}
  _{3}$) still manifest as continuous paths connecting correct and incorrect
  states across digit boundaries. However, the 2D projection exhibits noticeable
  entanglement, where different trajectories appear to cross or overlap. We
  attribute this to projection artifacts. The arithmetic manifold for 4-term addition
  possesses a higher intrinsic dimensionality required to orthogonally separate four
  carry states. When compressed into 2D, these dimensions collapse, creating
  visual occlusions.

  However, these apparent overlaps disappear in a 3D UMAP projection,
  % (see\texttt{html} files in the supplementary materials), 
  confirming that the fibers
  remain geometrically distinct in the high-dimensional representation space.

  \subsection{Geometry of Error Scenarios}

  The schematic in Figure \ref{fig:irst-num4} (Right) illustrates the logical
  structure of this expanded manifold. The expansion of the carry range from $\{0
  , 1, 2\}$ to $\{0, 1, 2, 3\}$ increases the number of bridges connecting adjacent
  digit basins.

  For any two adjacent digit anchors $k$ and $k+1$, there are now three distinct
  IRSTs that traverse the boundary, corresponding to three potential error scenarios:
  \begin{enumerate}
    \item \textbf{Low Carry Confusion ($0 \leftrightarrow 1$):} Occurs on
      trajectory $\mathcal{T}_{k}$. The model confuses state $(k, c=0)$ with $(k+
      1, c=1)$.

    \item \textbf{Mid Carry Confusion ($1 \leftrightarrow 2$):} Occurs on
      trajectory $\mathcal{T}_{k-1}$. The model confuses state $(k, c=1)$ with $(
      k+1, c=2)$.

    \item \textbf{High Carry Confusion ($2 \leftrightarrow 3$):} Occurs on
      trajectory $\mathcal{T}_{k-2}$. The model confuses state $(k, c=2)$ with $(
      k+1, c=3)$.
  \end{enumerate}

  This confirms that the Noisy Quantization Model scales with task complexity.
  As the number of operands increases, the robust plateau between quantization thresholds
  shrinks (packing more fibers into the same activation norm), statistically
  increasing the probability of geometric slippage (see \cref{fig:buthtub2}).

  \begin{algorithm}
    [ht]
    \caption{Inference-Time Correction via Dual-Stream Consistency}
    \label{alg:correction}
    \begin{algorithmic}
      [1] \STATE {\bfseries Input:} Activation $\boldsymbol{h}_{p}^{(L)}$,
      Current Prediction $\hat{s}_{p}$, Raw Sum Probe $f_{\theta_r}$, Potential
      Probe $f_{\theta_{\phi}}$, Tolerance $\delta$ \STATE {\bfseries Output:} Final
      Token $s_{final}$

      \STATE \COMMENT{\textit{1. Dual-Stream Decoding}} \STATE $\hat{r}_{p}\leftarrow
      f_{\theta_r}(\boldsymbol{h}_{p}^{(L)})$ \COMMENT{Decode Local Calculation Stream}
      \STATE
      $\hat{\Phi}_{p}\leftarrow f_{\theta_{\phi}}(\boldsymbol{h}_{p}^{(L)})$
      \COMMENT{Decode Global Context Stream}

      \STATE \COMMENT{\textit{2. Construct Plausible Carry Set}} \STATE $\mathcal{K}
      _{p}(\delta) \leftarrow \{ \lfloor \phi \rfloor \mid \phi \in [\hat{\Phi}_{p}
      - \delta, \hat{\Phi}_{p}+ \delta] \}$ \COMMENT{$\delta$-neighborhood of potential}

      \STATE \COMMENT{\textit{3. Consistency Verification}} \STATE
      $is\_consistent \leftarrow \text{False}$ \IF{$\exists c \in \mathcal{K}_{p}(\delta) \text{ s.t. }\hat{s}_{p}\equiv (\hat{r}_{p}+ c) \pmod{10}$}
      \STATE $is\_consistent \leftarrow \text{True}$ \ENDIF

      \STATE \COMMENT{\textit{4. Intervention Decision}} \IF{$is\_consistent$}
      \STATE $s_{final}\leftarrow \hat{s}_{p}$ \COMMENT{Preserve original output}
      \ELSE \STATE $s_{final}\leftarrow (\hat{r}_{p}+ \lfloor \hat{\Phi}_{p}\rfloor
      ) \pmod{10}$ \COMMENT{Intervene} \ENDIF

      \STATE \textbf{return} $s_{final}$
    \end{algorithmic}
  \end{algorithm}

  \section{Details for Inference-Time Correction Method}
  \label{app:algo}

  \subsection{Experiment Settings}
  We utilize Qwen3-4B, Qwen3-8B and Gemma-3-4B-IT to conduct the experiments. The
  dataset consists of $10,000$ addition problems involving 3 numbers, where each
  number is a positive integer with exactly 10 digits. We split the dataset into
  training, validation, and test sets with a ratio of $0.8 : 0.1 : 0.1$. To
  quantitatively evaluate the efficacy of our method, the main text focuses on three
  primary metrics: \textbf{Token Accuracy (Token Acc)}; \textbf{True Positive
  Correction (TP)}, which measures the proportion of originally incorrect tokens
  successfully rectified by the method; and \textbf{False Positive Preservation
  (FP)}, which measures the proportion of originally correct tokens that remain unchanged
  (following \citep{Sun25Probing}). The complete implementation details of our proposed
  inference-time correction protocol are summarized in \cref{alg:correction}.
  Unless otherwise noted, all probes described below are trained on the hidden states
  extracted from the final layer combing all positions. All experiments were
  conducted on a single NVIDIA GeForce RTX 3090 GPU. Our experiments were implemented
  using PyTorch \citep{paszke2019pytorch} and HuggingFace Transformers \citep{wolf2020transformers}.
  All models were run with \texttt{bfloat16} precision for efficient memory usage.

  \paragraph{Dual-Stream Probe}
  The raw sum probe $f_{\theta_r}$ is a classification MLP trained to minimize
  Cross-Entropy loss against the ground truth $r_{p}\pmod{10}$. The architecture
  consists of an input layer, two hidden layers of 512 and 128 dimensions
  respectively, and an output layer projecting to 10 classes. The carry
  potential probe $f_{\theta_{\phi}}$ is a regression MLP with a similar
  architecture but with a scalar output. It is trained via MSE loss to estimate the
  continuous potential $\Phi_{p}$. Both probes are optimized using the Adam
  optimizer with a learning rate of $10^{-3}$ and early stopping based on
  validation accuracy (for $f_{\theta_r}$) or validation MSE (for $f_{\theta_{\phi}}$).

  \paragraph{Replacement}
  This method utilizes an MLP (same architecture as $f_{\theta_r}$) to predict the
  ground truth digit and directly replaces the model's generation with the MLP's
  prediction during inference.

  \paragraph{Re-prompting \citep{Sun25Probing}}
  This method utilizes the probe as an error detector to trigger a natural
  language correction mechanism. The architecture of the probe is identical to that
  in the Replacement method. If the model's generated token matches the probe's
  prediction, it is appended to \texttt{\{current\_output\}}. If not, we append
  a specific correction suffix to the context: \textit{"That step looks incorrect.
  Let's re-do just this step: \{expression\} = \{current\_output\}"}. The model is
  then forced to regenerate the current digit based on this augmented context.
  Our evaluation reveals performance degradation compared to the baseline. We attribute
  this to KV-cache representation locking, where erroneous vector states persist
  and bias attention despite lexical corrections, and to asymmetric decision boundaries,
  which render high-confidence errors resistant to weak intervention signals
  while introducing new noise during regeneration.

  \paragraph{Steering}
  Steering guided by probe \citep{von2024language,taufeeque2024planning,chenglinearly,bhalla2024towards}
  is an emerging alternative technique. In this method, we first train a linear
  classification probe $P$. For a given hidden state $\boldsymbol{h}_{p}^{(L)}$,
  the probe predicts a target class
  $\hat{c}=\arg\max(W \boldsymbol{h}_{p}^{(L)})$, where $W$ is the weight matrix
  of $P$. The steering vector $\boldsymbol{v}$ is defined as the row in $W$
  corresponding to class $\hat{c}$ \citep{mallen2024eliciting,li2023inference}.
  The activation is then modified as
  ${\boldsymbol{h}'}_{p}^{(L)}= \boldsymbol{h}_{p}^{(L)}+ \lambda \cdot \boldsymbol
  {v}$. The steering strength $\lambda$ is selected via grid search over $\{0.0,
  0.25, 0.5, 0.75, 1.0\}$ to maximize sample-level accuracy on the validation set.

  \begin{figure}[tb]
    \centering
    \includegraphics[width=0.5\linewidth]{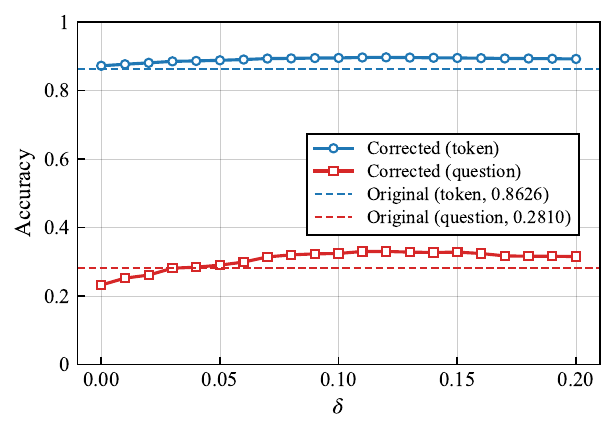}
    \vspace{-0.1in}
    \caption{\textbf{Impact of Tolerance $\delta$.} Token accuracy (left axis, blue)
    and Question accuracy (right axis, orange) under different values of $\delta$.
    The peak at $\delta \approx 0.12$ validates the existence of a noise margin
    in the model's carry potential estimation.}
    \label{fig:delta}
    \vspace{-0.1in}
  \end{figure}

  \subsection{Further Research on Tolerance}
  The set of plausible carries $\mathcal{K}_{p}(\delta)$ effectively defines a
  confidence interval in the carry space:
  \begin{equation}
    \mathcal{K}_{p}(\delta) = \left\{ \lfloor \phi \rfloor \mid \phi \in [\hat{\Phi}
    _{p}- \delta, \hat{\Phi}_{p}+ \delta] \right\},
  \end{equation}
  where $\delta$ is a hyperparameter representing tolerance.

  To investigate the influence of $\delta$ on both token and question accuracies,
  we conducted inference-time correction experiments by varying $\delta$ from 0
  to 0.2 in increments of 0.01. We evaluated performance using two metrics:
  \textbf{Token Accuracy}, which measures the proportion of individual digits
  correctly generated across all positions, and \textbf{Question Accuracy}, a stricter
  metric that counts a prediction as correct only if the entire generated number
  string perfectly matches the ground truth.

  As illustrated in Figure \ref{fig:delta}, both token and question accuracies
  exhibit a non-monotonic trend, initially ascending before subsequently declining.
  The metrics peak simultaneously at $\delta=0.12$, achieving a token accuracy of
  0.8973 and a question accuracy of 0.3300, while the minimum performance is
  recorded at $\delta=0$ (Token Acc: 0.8727, Ques Acc: 0.2320). Notably, the pronounced
  fluctuation observed in question accuracy aligns with the \textbf{Noisy
  Quantization Model}: strictly enforcing consistency ($\delta=0$) introduces false
  corrections near quantization boundaries, whereas a moderate tolerance ($\delta
  \approx 0.1$) effectively filters out this noise while correcting genuine
  geometric slippages.

  The set of plausible carries $\mathcal{K}_{p}(\delta)$ therefore functions as
  a narrow confidence interval in carry space, allowing the method to preserve
  ambiguous but still plausible internal states rather than forcing an unnecessarily
  sharp quantization decision.

  \begin{table}[!b]
    \caption{\textbf{Error-type analysis} of the Dual-Stream Consistency method
    on Qwen3-4B for 10-digit addition at position $p=4$.}
    \label{tab:error_type_matrix}
    \centering
    \begin{small}
      \begin{sc}
        \begin{tabular}{lccc}
          \toprule Error Type                & Fixed & Unfixed & Total \\
          \midrule Off-by-One ($\pm 1$)      & 96    & 561     & 657   \\
          Other Errors                       & 12    & 36      & 48    \\
          \midrule Total                     & 108   & 597     & 705   \\
          \bottomrule
        \end{tabular}
      \end{sc}
    \end{small}
    \vskip -0.1in
  \end{table}

  As shown in \cref{tab:error_type_matrix}, the vast majority (93.19\%) of the
  errors originally produced by the model are off-by-one errors. Among the errors
  successfully corrected by our method, the majority (88.9\%) are also off-by-one
  cases, confirming that the dual-stream intervention primarily targets the boundary-crossing
  failure mode emphasized throughout the paper.

  \begin{table*}[!t]
    \caption{Performance comparison on Qwen3-4b, Gemma-3-4b-it and Qwen3-8b using
    various inference-time correction methods.}
    \label{tab:generalize}
    \begin{center}
      \begin{small}
        \begin{sc}
          \begin{tabular}{lccccc}
            \toprule Method                                                                              & Token Acc       & Ques Acc        & Modified Rate & TP Correction & FP Preservation \\
            \midrule \multicolumn{6}{l}{\textit{Model: Qwen3-4B \quad Task: 10-digit addition (3 terms)}} \\
            \midrule Original                                                                            & 0.8626          & 0.2810          & /             & /             & /               \\
            Steering                                                                                     & 0.8827          & 0.2820          & 0.0775        & 0.3058        & 0.9697          \\
            Replacement                                                                                  & 0.8913          & 0.3000          & 0.0710        & 0.3173        & 0.9765          \\
            Re-Prompting                                                                                 & 0.7990          & 0.2610          & 0.1040        & 0.0008        & 0.9998          \\
            Dual-Stream Probe ($\delta=0$)                                                               & 0.8727          & 0.2320          & 0.1240        & 0.4439        & 0.9407          \\
            Dual-Stream Probe ($\delta=0.05$)                                                            & 0.8887          & 0.2900          & 0.0844        & 0.3551        & 0.9675          \\
            Dual-Stream Probe ($\delta=0.1$)                                                             & \textbf{0.8956} & \textbf{0.3240} & 0.0646        & 0.3046        & 0.9813          \\
            \midrule                                                                                      % ---原有的第二组数据: Gemma-3-4b-it ---
            \multicolumn{6}{l}{\textit{Model: Gemma-3-4B-IT \quad Task: 10-digit addition (3 terms)}}     \\
            \midrule Original                                                                            & 0.7286          & 0.3170          & /             & /             & /               \\
            Steering                                                                                     & 0.8210          & 0.3010          & 0.1417        & 0.3319        & 0.9531          \\
            Replacement                                                                                  & 0.8499          & \textbf{0.3780} & 0.1085        & 0.3405        & 0.9732          \\
            Re-Prompting                                                                                 & 0.7141          & 0.3210          & 0.1065        & 0.1757        & 0.9133          \\
            Dual-Stream Probe ($\delta=0$)                                                               & 0.8586          & 0.3470          & 0.1297        & 0.4046        & 0.9643          \\
            Dual-Stream Probe ($\delta=0.05$)                                                            & \textbf{0.8598} & 0.3730          & 0.0967        & 0.3264        & 0.9837          \\
            Dual-Stream Probe ($\delta=0.1$)                                                             & 0.8492          & 0.3580          & 0.0841        & 0.2726        & 0.9897          \\
            \midrule                                                                                      % --- 原有的第三组数据: Qwen3-8b ---
            \multicolumn{6}{l}{\textit{Model: Qwen3-8B \quad Task: 12-digit addition (3 terms)}}          \\
            \midrule Original                                                                            & 0.9329          & 0.5630          & /             & /             & /               \\
            Steering                                                                                     & 0.9470          & 0.5770          & 0.0784        & 0.3108        & 0.9688          \\
            Replacement                                                                                  & \textbf{0.9547} & \textbf{0.6440} & 0.0391        & 0.4222        & 0.9891          \\
            Re-Prompting                                                                                 & 0.8806          & 0.5360          & 0.0738        & 0.0843        & 0.9379          \\
            Dual-Stream Probe ($\delta=0$)                                                               & 0.9187          & 0.4390          & 0.0782        & 0.4439        & 0.9688          \\
            Dual-Stream Probe ($\delta=0.05$)                                                            & 0.9378          & 0.5350          & 0.0425        & 0.3315        & 0.9838          \\
            Dual-Stream Probe ($\delta=0.1$)                                                             & 0.9462          & 0.5710          & 0.0266        & 0.2638        & 0.9933          \\
            \bottomrule
          \end{tabular}
        \end{sc}
      \end{small}
    \end{center}
    \vskip -0.1in
  \end{table*}

  \subsection{Generalization Analysis}
  To assess the robustness of our approach across different architectures and task
  complexities, we extend our evaluation to Gemma-3-4B-IT (10-digit addition)
  and Qwen3-8B (12-digit addition), as detailed in Table \ref{tab:generalize}.

  In this broader analysis, we introduce two additional metrics to provide a
  holistic view of performance: \textbf{Question Accuracy (Ques Acc)} as
  explained above, and \textbf{Modified Rate}, which quantifies the frequency with
  which the inference-time method alters the original model output.

  As shown in Table \ref{tab:generalize}, the Dual-Stream Probe maintains
  competitive performance. We note that in some cases (particularly Qwen3-8B),
  the Dual-Stream Probe yields slightly lower accuracy metrics compared to the
  strict Replacement method. Crucially, these results serve as a causal
  validation of the IRST geometry rather than a mere performance boost. The ability
  to correct errors using only internal consistency (Dual-Stream) implies that the
  model retains correct latent information even when the output quantization fails.
  The Replacement method, while effective, relies on the probe's external
  supervision; the Dual-Stream method relies on the model's own internal coherence.

  \section{Layer-Wise Analysis of Internal Activations}
  \label{app:probe}
  \subsection{Probe Analysis}
  To understand where and how the arithmetic logic emerges within the model , we
  conducted a comprehensive layer-wise probing analysis using Qwen3-4B. We use the
  same dataset and dataset split ratio as in Appendix \ref{app:algo}. We trained
  separate MLP probes on the hidden states of each layer (0–36) on all positions
  to predict six target variables which are consistent with the main text. The specific
  definitions of these targets are as follows:

  \textbf{Correctness} A binary label indicating whether the model's generated
  digit successfully matches the ground truth.

  \textbf{Carry Potential} A continuous regression target $\Phi_{p}$, described
  in Section \ref{subsec:carry_potential}.For evaluation and testing, the floor function
  is applied to interpret this continuous prediction as a discrete integer carry
  value.

  \textbf{Ground Truth} The mathematically correct digit expected at the current
  position (integer label 0–9).

  \textbf{Input Carry} The actual integer carry value transmitted from the previous
  column (integer label 0-2 in our specific task).

  \textbf{Model Output} The specific digit token actually generated by the model
  (integer label 0–9).

  \textbf{Raw Sum} The local sum of the operand digits in the current column modulo
  10, excluding any incoming carry (integer label 0–9).

\begin{figure*}[tb] % 使用 figure* 跨栏
    \centering
    % --- 第一行 (Top Row) ---
    \begin{subfigure}
      [b]{0.32\textwidth}
      \centering
      \includegraphics[width=\linewidth]{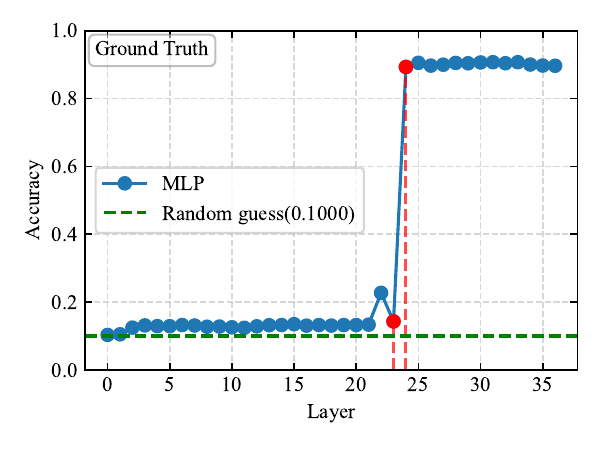} % 替换文件名
      \caption{Ground Truth Probe}
      \label{fig:row1_1}
    \end{subfigure}
    \hfill % 弹性间距
    \begin{subfigure}
      [b]{0.32\textwidth}
      \centering
      \includegraphics[width=\linewidth]{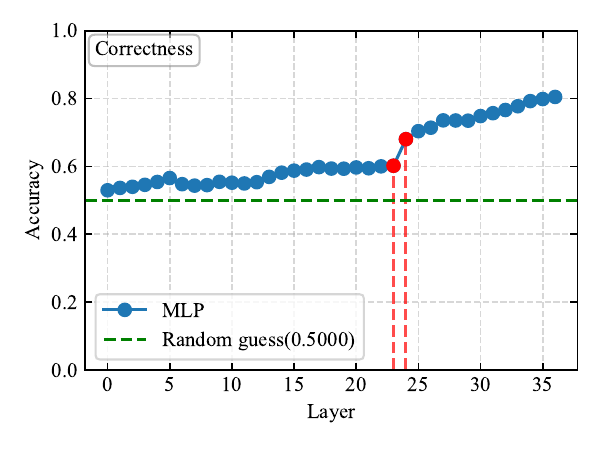} % 替换文件名
      \caption{Correctness Probe}
      \label{fig:row1_2}
    \end{subfigure}
    \hfill % 弹性间距
    \begin{subfigure}
      [b]{0.32\textwidth}
      \centering
      \includegraphics[width=\linewidth]{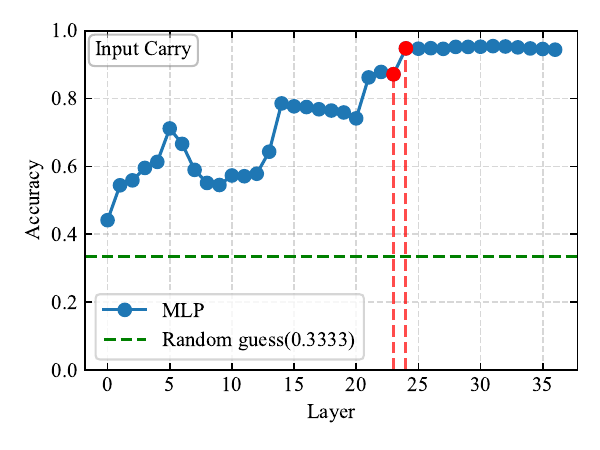} % 替换文件名
      \caption{Input Carry Probe}
      \label{fig:row1_3}
    \end{subfigure}

    % --- 行间距 (根据审美调整，通常 0.1in 到 0.2in) ---
    \vspace{0.15in}

    % --- 第二行 (Bottom Row) ---
    \begin{subfigure}
      [b]{0.32\textwidth}
      \centering
      \includegraphics[width=\linewidth]{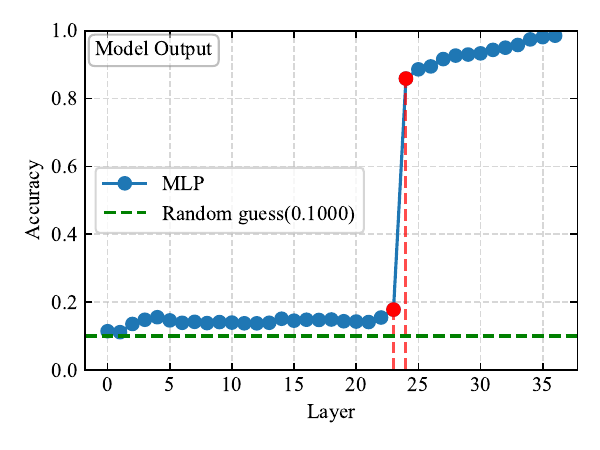} % 文件名
      \caption{Model Output Probe}
      \label{fig:row2_1}
    \end{subfigure}
    \hfill % 弹性间距
    \begin{subfigure}
      [b]{0.32\textwidth}
      \centering
      \includegraphics[width=\linewidth]{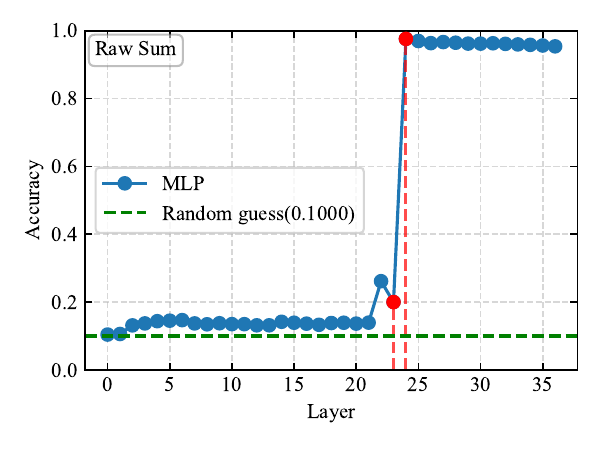} % 文件名
      \caption{Raw Sum Probe}
      \label{fig:row2_2}
    \end{subfigure}
    \hfill % 弹性间距
    \begin{subfigure}
      [b]{0.32\textwidth}
      \centering
      \includegraphics[width=\linewidth]{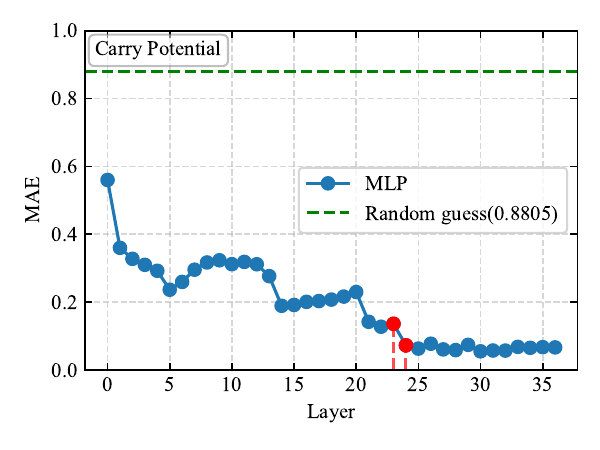} % 文件名
      \caption{Carry Potential Probe}
      \label{fig:row2_3}
    \end{subfigure}

    % --- 格式要求：Caption 前留 0.1 inches ---
    \vspace{0.1in}

    % --- 格式要求：9 point type (用 small 环境模拟) ---
    \begin{small}
      \caption{Layer-wise performance of different probes trained on Qwen3-4B
      combing all positions.}
      \label{fig:layerprobe}
    \end{small}

    % --- 格式要求：Caption 后留 0.1 inches ---
    \vspace{0.1in}
  \end{figure*}

  From the results in Figure \ref{fig:layerprobe}, across all arithmetic-related
  variables (Raw Sum, Ground Truth, Model Output), we observe a distinct step-function
  behavior. The information regarding the specific digit values remains nearly imperceptible
  (accuracy close to random guessing) throughout the early and middle layers. A
  sudden informational surge occurs at Layer 24. The regression error for the
  continuous carry potential remains high in early layers but drops at Layer 24,
  coinciding with the emergence of the Raw Sum. The input carry information
  shows a more volatile trajectory in intermediate layers before stabilizing and
  peaking alongside the other metrics. This synchronization supports our Dual-Stream
  hypothesis: the model appears to disentangle and resolve both the local
  arithmetic facts (Raw Sum) and the global context (Carry) simultaneously to
  synthesize the final output.

  \begin{figure}[t]
    \centering
    \centerline{\includegraphics[width=0.82\columnwidth]{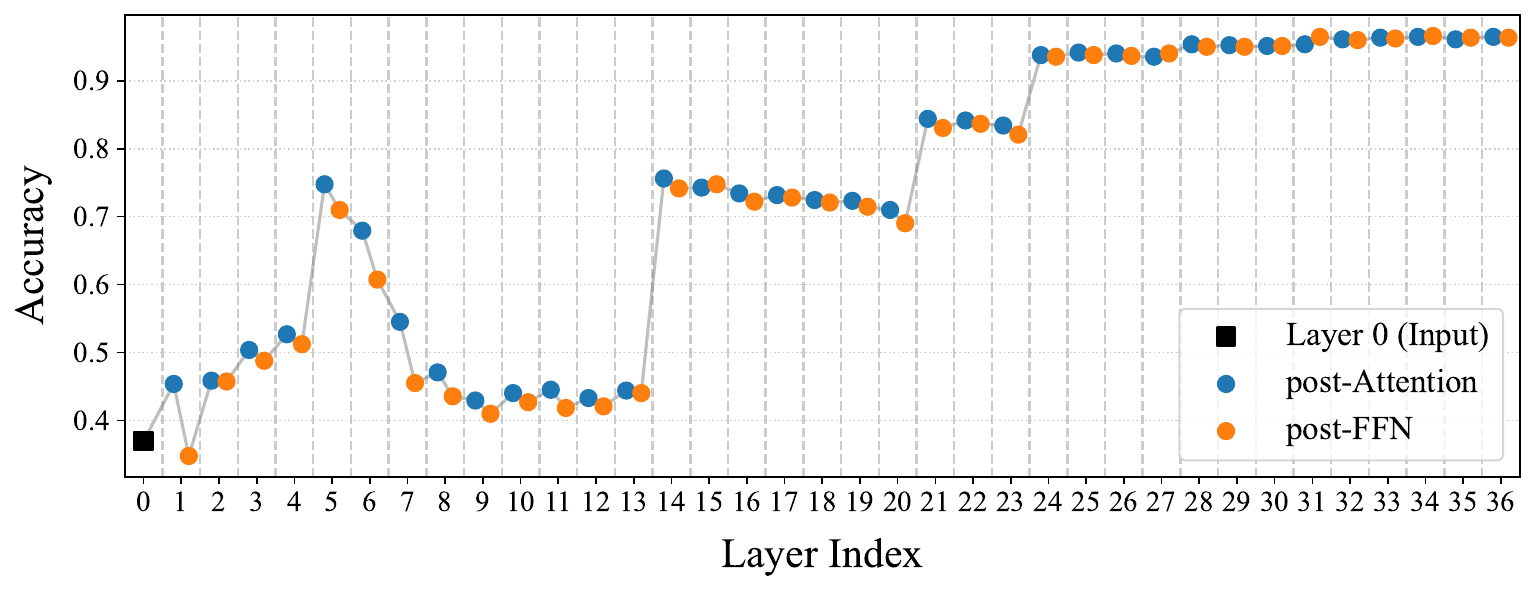}}
    \caption{\textbf{Layer-wise input-carry decoding accuracy for attention and
    FFN outputs.} Attention blocks exhibit sharper stepwise gains, while FFN outputs
    largely follow these updates, suggesting that carry information is consolidated
    through a staged pipeline across layers.}
    \label{fig:attn_ffn_carry} \vskip -0.1in
  \end{figure}

  To partially localize where the carry signal is written, we separately probe the
  outputs of the attention and FFN blocks. As shown in \cref{fig:attn_ffn_carry},
  the attention modules exhibit the sharpest stepwise gains, whereas the FFN
  outputs mostly lag behind and mirror these updates. This pattern suggests that
  carry information is first consolidated by attention and then propagated through
  a staged feed-forward refinement process.

  \subsection{UMAP Visualization}
  To further interpret the dynamic evolution of representations, we employ UMAP
  to perform dimensionality reduction on the hidden states of intermediate
  layers. Using the same Qwen3-4B model and the dataset described in Appendix \ref{app:algo},
  we apply \textbf{Aligned UMAP} across all layers. Unlike standard UMAP, this technique
  enforces temporal consistency by regularizing the projection of each layer based
  on the geometry of the preceding layer. The results are visualized in Figure
  \ref{fig:layer_umap_3x3}, and the complete evolutionary dynamics are provided as
  \texttt{GIF} files in the supplementary materials.

  The results reveal that from Layers 1 to 23 (Figure \ref{fig:layer_umap_3}-\ref{fig:layer_umap_23}),
  samples exhibit a persistent $10\times 10$ hierarchical clustering structure,
  with ten macro-clusters representing the previously generated digit (Position 3),
  each contain ten micro-clusters. This structured organization remains stable
  throughout the early layers, with only minor shifts and local expansions or contractions.
  The precise mechanism of micro-clusters remains to be fully elucidated; we hypothesize
  they may represent the residual encoding of previously generated digits (e.g.,
  the value at Position 2) or latent carry signals.

  Consistent with the sudden informational surge observed in our probing experiments,
  a dramatic mutation occurs at Layer 24 (Figure \ref{fig:layer_umap_24}), where
  the hierarchical clusters collapse into an elliptical manifold. In the late
  layers (Figure \ref{fig:layer_umap_30}-\ref{fig:layer_umap_36}), this manifold
  gradually dissociates into distinct IRST curves. This transition marks the
  point where the model resolves the latent variables in explicit arithmetic
  logic.

  Notably, the IRSTs in Figure \ref{fig:layer_umap_36} appear disjoint,
  contrasting with the continuous manifold in Figure \ref{fig:umap1}. This fragmentation
  is an artifact of Aligned UMAP: by penalizing large displacements to ensure
  temporal continuity, the distinct clustering of early layers acts as a geometric
  anchor that visually tears the emerging manifold. Thus, while Figure \ref{fig:umap1}
  depicts the intrinsic geometry, Figure \ref{fig:layer_umap_36} captures the
  path-dependent evolution.

  \begin{figure}[tb]
    \centering
    % 第一行
    \begin{subfigure}
      [b]{0.3\textwidth}
      \centering
      \includegraphics[width=\textwidth]{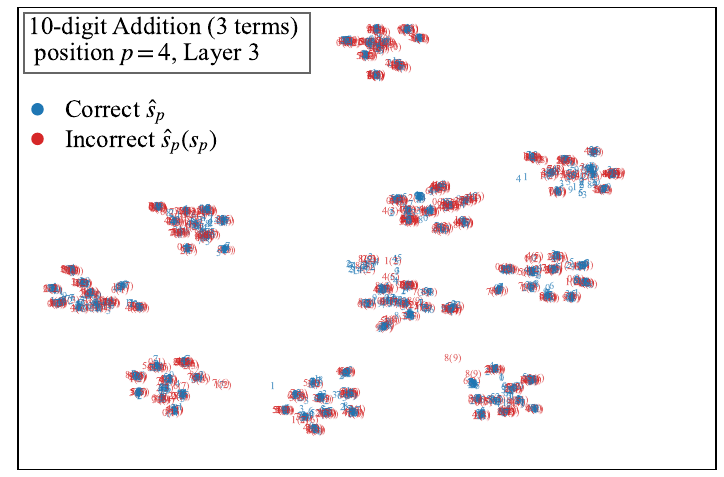}
      \caption{Layer 3}
      \label{fig:layer_umap_3}
    \end{subfigure}
    \hfill
    \begin{subfigure}
      [b]{0.3\textwidth}
      \centering
      \includegraphics[width=\textwidth]{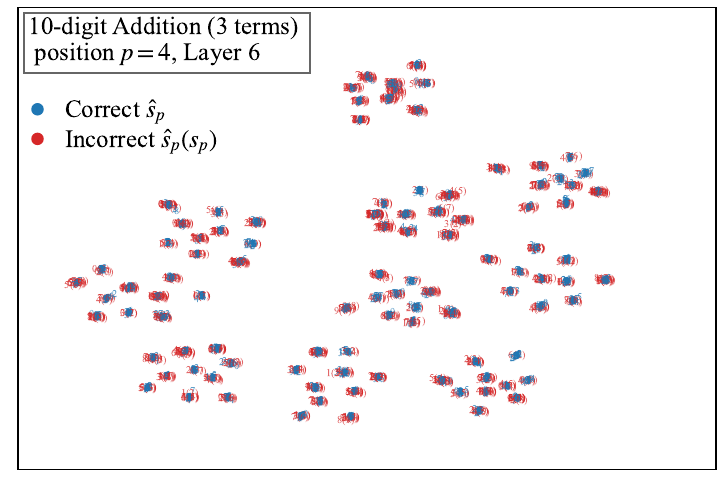}
      \caption{Layer 6}
      \label{fig:layer_umap_6}
    \end{subfigure}
    \hfill
    \begin{subfigure}
      [b]{0.3\textwidth}
      \centering
      \includegraphics[width=\textwidth]{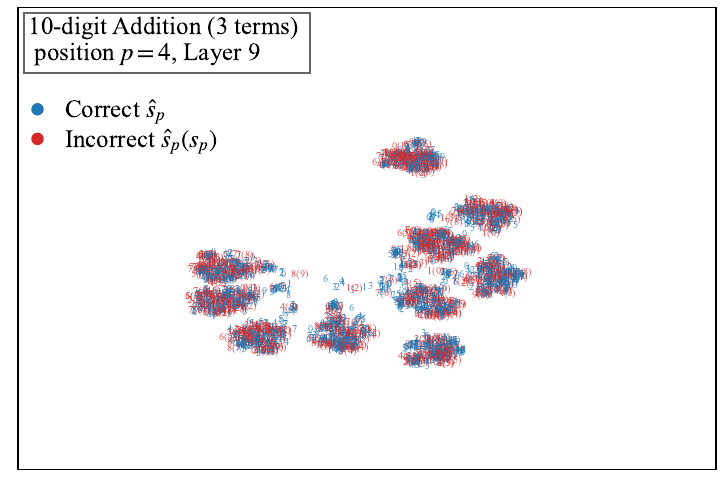}
      \caption{Layer 9}
      \label{fig:layer_umap_9}
    \end{subfigure}

    \vspace{10pt} % 行间距

    % 第二行
    \begin{subfigure}
      [b]{0.3\textwidth}
      \centering
      \includegraphics[width=\textwidth]{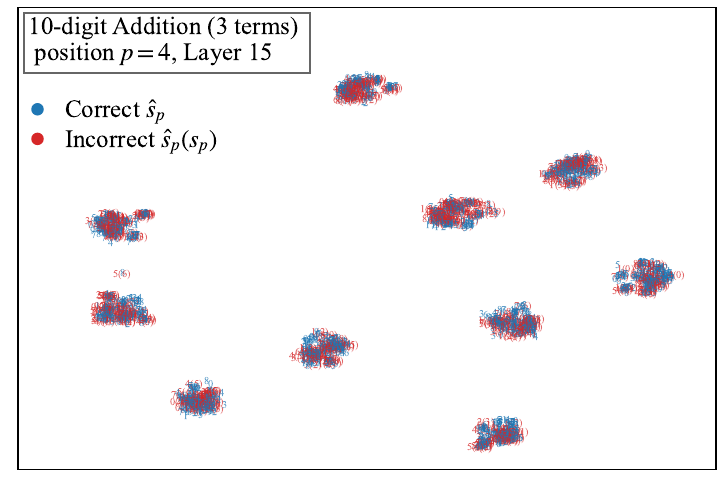}
      \caption{Layer 15}
      \label{fig:layer_umap_15}
    \end{subfigure}
    \hfill
    \begin{subfigure}
      [b]{0.3\textwidth}
      \centering
      \includegraphics[width=\textwidth]{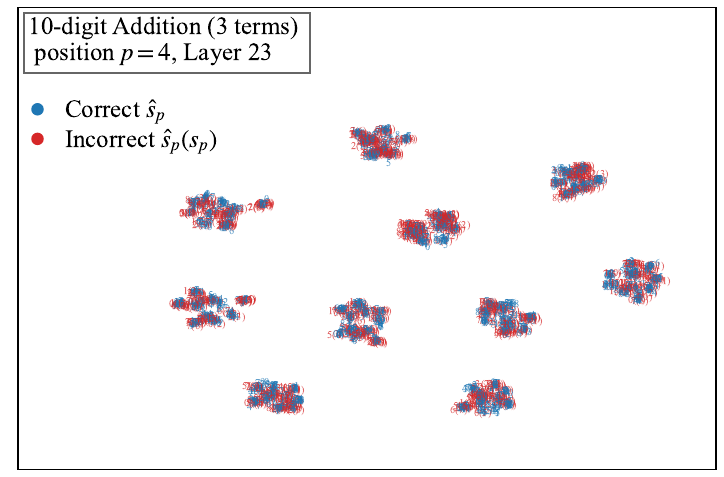}
      \caption{Layer 23}
      \label{fig:layer_umap_23}
    \end{subfigure}
    \hfill
    \begin{subfigure}
      [b]{0.3\textwidth}
      \centering
      \includegraphics[width=\textwidth]{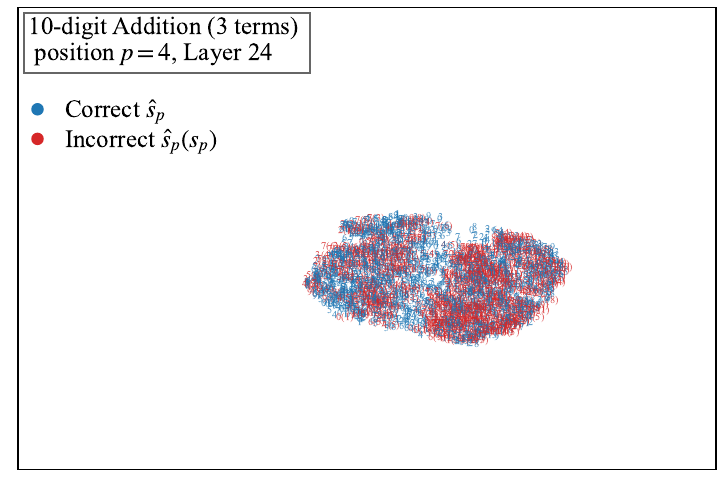}
      \caption{Layer 24}
      \label{fig:layer_umap_24}
    \end{subfigure}

    \vspace{10pt} % 行间距

    % 第三行
    \begin{subfigure}
      [b]{0.3\textwidth}
      \centering
      \includegraphics[width=\textwidth]{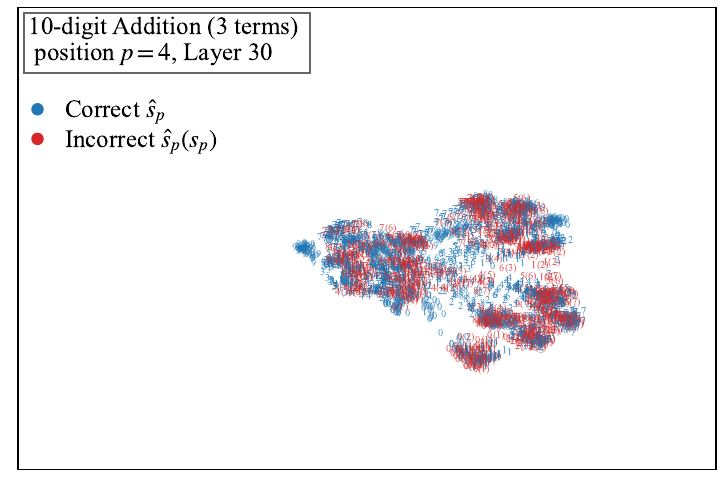}
      \caption{Layer 30}
      \label{fig:layer_umap_30}
    \end{subfigure}
    \hfill
    \begin{subfigure}
      [b]{0.3\textwidth}
      \centering
      \includegraphics[width=\textwidth]{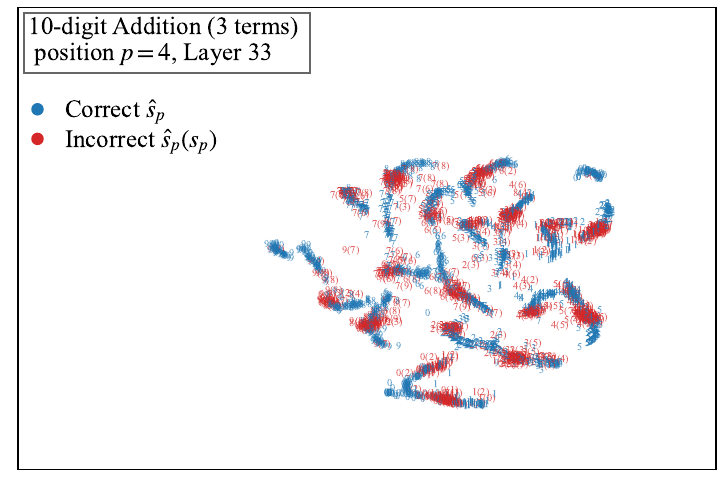}
      \caption{Layer 33}
      \label{fig:layer_umap_33}
    \end{subfigure}
    \hfill
    \begin{subfigure}
      [b]{0.3\textwidth}
      \centering
      \includegraphics[width=\textwidth]{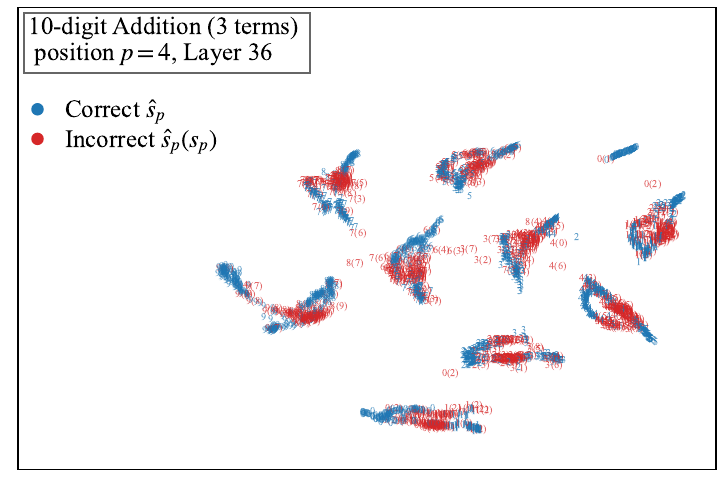}
      \caption{Layer 36}
      \label{fig:layer_umap_36}
    \end{subfigure}

    \caption{Layer-wise Alighed UMAP visualization.}
    \label{fig:layer_umap_3x3}
  \end{figure}

  \section{Probe vs. Logit Lens: Evidence of Latent Arithmetic States}
  \label{app:logit_lens}

  To further validate that the geometric structures (IRSTs) captured by our
  probes represent intrinsic internal computations rather than trivial output
  reflections, we compared the performance of our Linear Probes against the \textbf{Logit
  Lens} technique \citep{nostalgebraist2020logitlens}.

\begin{figure*}[ht]
    \centering
    \begin{minipage}[b]{0.45\textwidth}
      \includegraphics[width=\linewidth]{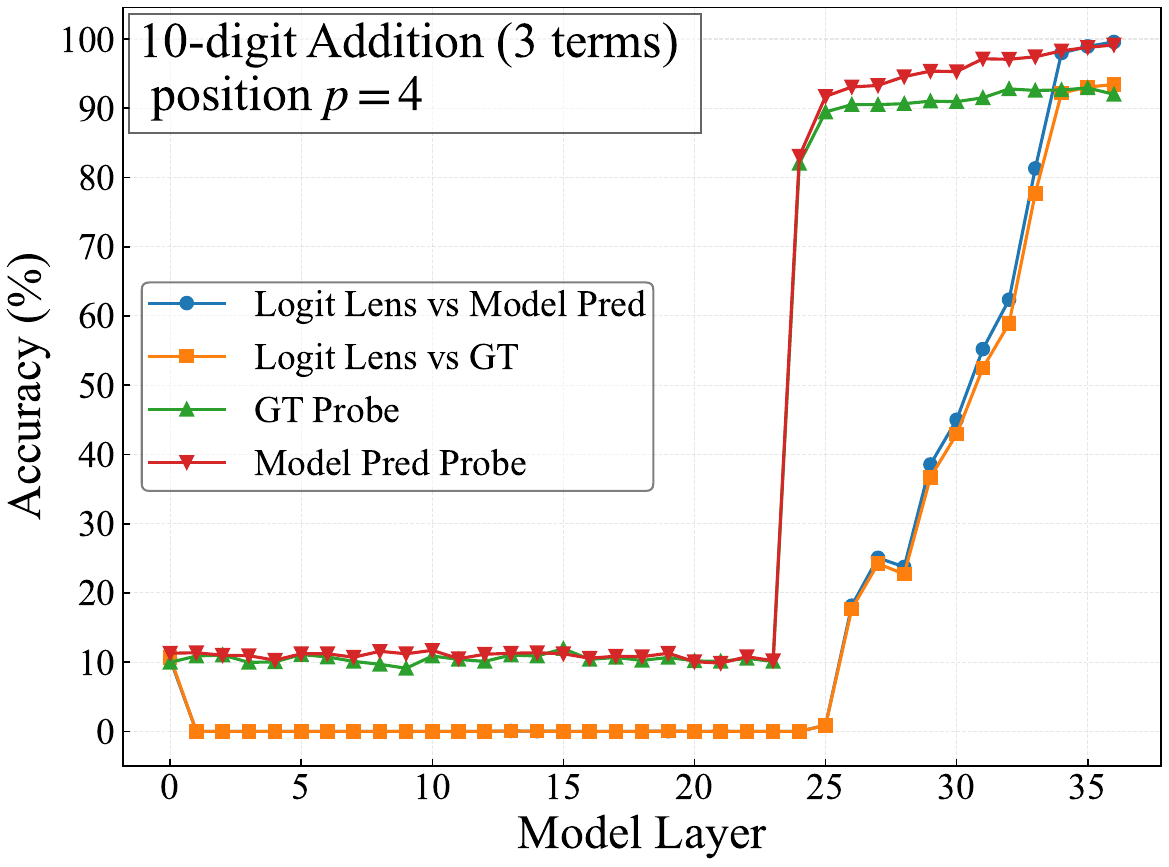}
    \end{minipage}
    \hfill
    \begin{minipage}[b]{0.45\textwidth}
      \includegraphics[width=\linewidth]{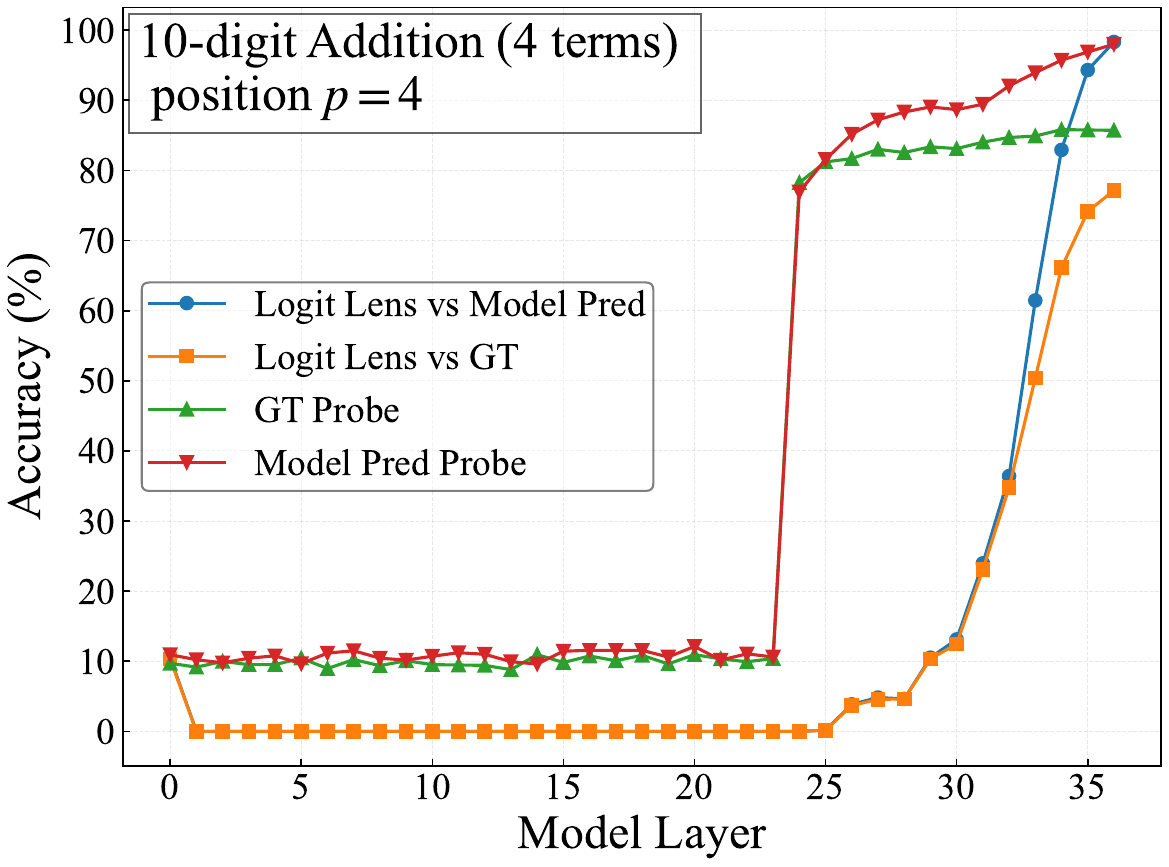}
    \end{minipage}
    \caption{\textbf{Layer-wise Performance Comparison: Linear Probes vs. Logit
    Lens.} The Green and Red lines represent the accuracy of linear probes trained
    on $\boldsymbol{h}_{p}^{(l)}$ for the Ground Truth (GT) and Model Prediction
    (Pred), respectively. The Blue and Orange lines represent the accuracy of the
    Logit Lens (applying the unembedding matrix directly). \textbf{Key
    Observation:} There is a significant \textit{decoding lag}. Probes successfully
    decode the arithmetic state ($\text{Accuracy}> 80\%$) starting around Layer 24,
    whereas the Logit Lens fails to extract meaningful information until Layer
    30 (the ``readout phase''). This gap indicates that the arithmetic computation
    occurs in a latent geometric subspace (the IRST) that is linearly separable but
    not yet aligned with the output vocabulary.}
    \label{fig:logit_lens_comparison}
  \end{figure*}

  The Logit Lens applies the final layer's unembedding matrix $W_{U}$ to intermediate
  hidden states $\boldsymbol{h}_{p}^{(l)}$ to project them directly into the vocabulary
  space. We evaluated both methods on identifying the Ground Truth Digit (GT)
  and the Model's Final Output Digit (Pred) across all 36 layers. The results for
  3-term and 4-term addition are shown in Figure \ref{fig:logit_lens_comparison}.

  \subsection{Analysis of the Decoding Gap}

  The comparison reveals three distinct phases of processing:
  \begin{enumerate}
    \item \textbf{Preparation Phase (Layers 0--23):} Both probes and Logit Lens
      show near-random accuracy ($< 10\%$), suggesting that the specific digit information
      for the current position is not yet fully localized or linearly decodable.

    \item \textbf{Latent Computation Phase (Layers 24--29):} This is the
      critical window where our geometric analysis focuses. Linear probes achieve
      high accuracy ($>90\%$), indicating that the model has internally resolved
      the arithmetic state (Raw Sum and Carry). However, the Logit Lens accuracy
      remains low. This discrepancy proves that the internal representation at this
      stage is \textit{orthogonal} or \textit{unaligned} with the static word embedding
      space. The IRST structure exists here as an abstract mathematical
      representation, decoupled from the specific token IDs.

    \item \textbf{Readout Phase (Layers 30--36):} The Logit Lens accuracy
      sharply rises to match the probes, indicating that the internal state is finally
      rotated into the vocabulary space for token generation.
  \end{enumerate}

  This comparison strongly justifies the necessity of training probes (and
  analyzing the resulting geometry) rather than relying on the Logit Lens. The
  Logit Lens would miss the crucial ``Latent Computation Phase'' where the core
  arithmetic logic—and the geometric slippages we describe—actually unfold.

  %%%%%%%%%%%%%%%%%%%%%%%%%%%%%%%%%%%%%%%%%%%%%%%%%%%%%%%%%%%%%%%%%%%%%%%%%%%%%%%
  %%%%%%%%%%%%%%%%%%%%%%%%%%%%%%%%%%%%%%%%%%%%%%%%%%%%%%%%%%%%%%%%%%%%%%%%%%%%%%%
\end{document}